\documentclass{article}

\usepackage[final]{neurips_2025}

\usepackage[utf8]{inputenc} %
\usepackage[T1]{fontenc}    %

\usepackage{times}
\usepackage{epsfig}
\usepackage{graphicx}
\usepackage{url}            %
\usepackage{booktabs}       %
\usepackage{amsfonts}       %
\usepackage{nicefrac}       %
\usepackage{microtype}      %
\usepackage{color}
\usepackage{colortbl}
\usepackage{dsfont}
\usepackage{algorithm,algpseudocode}
\usepackage{algorithmicx}
\usepackage{amsmath,amssymb}
\usepackage{multirow}
\usepackage{makecell}
\usepackage{listings}
\usepackage{tabularx}
\usepackage{xcolor}
\usepackage[accsupp]{axessibility}
\newcounter{abs}

\usepackage{mathtools}
\usepackage{amsthm}
\usepackage{caption}
\algblock{ParFor}{EndParFor}
\algnewcommand\algorithmicparfor{\textbf{parfor}}
\algnewcommand\algorithmicpardo{\textbf{do}}
\algnewcommand\algorithmicendparfor{\textbf{end\ parfor}}
\algrenewtext{ParFor}[1]{\algorithmicparfor\ #1\ \algorithmicpardo}
\algrenewtext{EndParFor}{\algorithmicendparfor}
\usepackage{wrapfig}

\definecolor{Green}{HTML}{E5F8F6}
\definecolor{Blue}{HTML}{DAE3F5}
\definecolor{Purple}{HTML}{ECE3EC}
\definecolor{RPurple}{HTML}{FED8B1}
\definecolor{Yellow}{HTML}{fdffb7}
\definecolor{Grey}{rgb}{0.81176471, 0.81176471, 0.81176471}
\definecolor{DGreen}{rgb}{0.15882353, 0.80980392, 0.14705882}
\definecolor{ngreen}{HTML}{76B900}

\newcommand{\customfootnotetext}[2]{{%
\renewcommand{\thefootnote}{#1}%
\footnotetext[0]{#2}}}%
\usepackage{hyperref}
\usepackage{cleveref}

\usepackage{subfiles}  %

\title{GSPN-2: Efficient Parallel Sequence Modeling}

\author{Hongjun Wang\textsuperscript{1,2,$\dagger$}, Yitong Jiang$^1$, Collin McCarthy$^1$, David Wehr$^1$, Hanrong Ye$^1$, Xinhao Li$^3$,
\AND Ka Chun Cheung$^1$, Wonmin Byeon$^1$, Jinwei Gu$^1$, Ke Chen$^1$, Kai Han$^2$\thanks{Corresponding author.},
\AND Hongxu Yin$^1$, Pavlo Molchanov$^1$, Jan Kautz$^1$, Sifei Liu$^1$\\
$\textsuperscript{\rm 1}$NVIDIA \quad$\textsuperscript{\rm 2}$The University of Hong Kong  \quad$\textsuperscript{\rm 3}$University of California, San Diego
}

\begin{document}

\maketitle
\customfootnotetext{}{$\dagger$ Hongjun Wang was an intern at NVIDIA during the project.}

\begin{abstract}

Efficient vision transformer remains a bottleneck for high-resolution images and long-video related real-world applications. Generalized Spatial Propagation Network (GSPN) \cite{wang2025parallel} addresses this by replacing quadratic self-attention with a line-scan propagation scheme, bringing the cost close to linear in the number of rows or columns, while retaining accuracy. Despite this advancement, the existing GSPN implementation still suffers from (i) heavy overhead due to repeatedly launching GPU kernels, (ii) excessive data transfers from global GPU memory, and (iii) redundant computations caused by maintaining separate propagation weights for each channel. We introduce GSPN-2, a joint algorithm–system redesign. In particular, we eliminate thousands of micro-launches from the previous implementation into one single 2D kernel, explicitly pin one warp to each channel slice, and stage the previous column's activations in shared memory. On the model side, we introduce a compact channel propagation strategy that replaces per-channel matrices, trimming parameters, and align naturally with the affinity map used in transformer attention. 
Experiments demonstrate GSPN-2's effectiveness across image classification and text-to-image synthesis tasks, matching transformer-level accuracy with significantly lower computational cost. GSPN-2 establishes a new efficiency frontier for modeling global spatial context in vision applications through its unique combination of structured matrix transformations and GPU-optimized implementation. Project page: \url{https://whj363636.github.io/GSPN2/}    
\end{abstract}

\section{Introduction}
\label{sec:intro}

Vision transformers have underpinned nearly every state-of-the-art (SOTA) vision foundation model: text-to-image diffusion networks (e.g., Stable Diffusion~\cite{rombach2022high}), vision-language aligners such as CLIP~\cite{radford2021learning} and SigLIP~\cite{zhai2023sigmoid}, and modern detection/segmentation pipelines~\cite{liu2024grounding, kirillov2023segment} -- all depend on their dense, token-wise attention to encode visual concepts. Since this attention operator scales quadratically with the number of pixels, practical deployments still cap the input—SigLIP~\cite{zhai2023sigmoid}, for instance, limits the input images to $512\times 512$—to avoid prohibitive latency and memory. Recently, several efficient-attention variants have been proposed, such as FlashAttention~\cite{dao2022flashattention, dao2023flashattention2}, linear attention~\cite{rabe2021self,peng2021random,han2023flatten}, and state-space models~\cite{gu2023mamba,dao2024transformers}. Among them, Generalized Spatial Propagation Networks (GSPN)~\cite{wang2025parallel} uniquely replace 2D self-attention with a line-scan approach, which reduces the computational complexity from quadratic to approximately linear to the image's width or height. Remarkably, GSPN maintains or even surpasses baseline accuracy while achieving up to an 84× speed-up for 16 K-resolution diffusion inference.

While most efficient attention backends can reuse existing matrix-multiply or scan primitives, GSPN's line scan demands a completely new CUDA implementation. Standard Softmax attention breaks down into a series of GEMM-based matrix multiplies plus a softmax~\cite{shah2024flashattention}. FlashAttention~\cite{dao2022flashattention, dao2023flashattention2} instead fuses those steps into a single tiled GEMM loop. State-space methods like Mamba~\cite{gu2023mamba,dao2024transformers} recasts attention into a streaming recurrence and implement it with fast prefix-sum scans across tokens.

By contrast, GSPN adopts a 3-neighbour line-scan approach, which is neither a matrix multiply, nor a prefix scan—its dependency pattern would explode combinatorially. Therefore, a specifically built CUDA kernel for GSPN is required to unlock its sub-quadratic cost. %

\begin{figure}[h]
    \centering
    \includegraphics[width=1.0\textwidth]{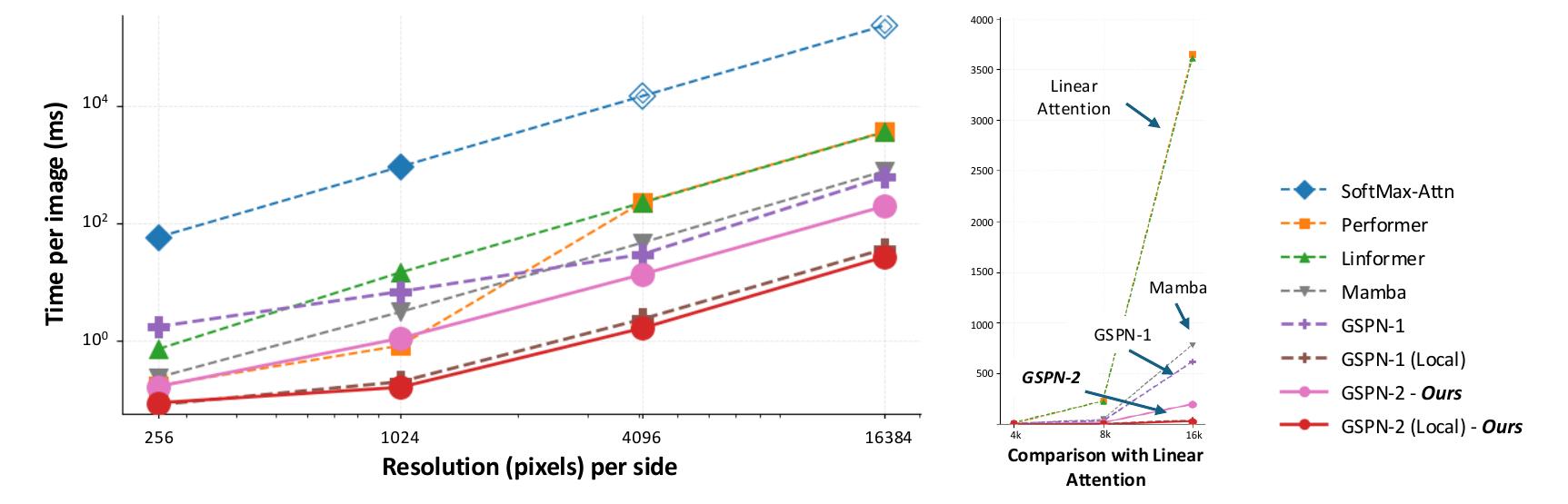}
    \caption{GSPN-2 achieves transformative performance improvements over GSPN-1\cite{wang2025parallel} and other efficient attention variants, running up to 30-50× faster across diverse input configurations on modern GPU architectures.}
    \label{fig:intro}
\end{figure}

While GSPN-1 offers theoretical advantages in computational complexity, its initial CUDA implementation \cite{wang2025parallel} struggled to translate these into practical speedups. 
Profiling reveals that GSPN-1’s reference CUDA code, which launches a tiny kernel for each column step, severely underutilizes the GPU, as it achieves just 3–8\% of peak memory bandwidth and low SM occupancy. 
This inefficiency stemmed from several critical bottlenecks:
(1) the kernel-launch overhead from thousands of separate launches, which prevents the SMs from staying fully busy;
(2) inefficient global‐memory (HBM) access, with each step reloading data without on-chip reuse or coalescing; and
(3) poor cache locality and growing runtime as channel counts increase.

To overcome the limitations of the original GSPN implementation, we introduce GSPN-2, an integrated algorithmic and kernel-level redesign. Specifically, GSPN-2 (a) consolidates all propagation steps into a single unified CUDA kernel, eliminating costly repeated launch overheads; (b) introduces a compact multi-channel propagation mechanism that projects features into a lower-dimensional proxy space to reduce concurrent thread blocks and maintain constant-time performance; and
(c) refines the grid and block configuration to improve warp-level efficiency and memory coalescing, with optional on-chip caching for reuse of hidden states. 
To further address the GPU concurrency bottleneck—where runtime increases sharply once the number of active thread blocks exceeds the device’s scheduling capacity—GSPN-2 employs a lightweight channel-projection strategy. By projecting the input tensor into a compact proxy space before propagation, the system reduces the effective computational dimension, improving cache reuse and maintaining high throughput even under large batch and channel counts (see~\Cref{sec:gspnv2_theory}).
These combined algorithmic and kernel-level improvements deliver substantial performance gains: on an NVIDIA A100 GPU, the runtime for a $1024 \times 1024 \times 8$ input decreases from 71.4 ms in GSPN-1 to just 1.8 ms in GSPN-2—achieving an overall 40× speedup (see Figure~\ref{fig:journey}).

Our experimental evaluation comprehensively validates GSPN-2. Rigorous efficiency analysis demonstrates that GSPN-2 runs up to $30\times$ faster than GSPN-1 across diverse input configurations, with performance profiling confirming near-optimal hardware utilization (over 90\% of theoretical peak memory bandwidth). We then validate GSPN-2's effectiveness across vision tasks: image classification and text-to-image synthesis. On ImageNet, GSPN-2 achieves accuracy comparable to transformer models at significantly lower computational cost. In text-to-image synthesis, GSPN-2 significantly improves semantic consistency and visual quality when integrated with existing diffusion models. These results confirm GSPN-2 as a versatile component for efficiently modeling global spatial context across diverse vision applications.

\section{Related Works}
\label{app:rw}

\noindent\textbf{Efficient Attention Mechanisms.} Transformer architectures~\cite{vaswani2017attention} have become foundational components in modern vision and language models, but their quadratic computational complexity with respect to sequence length creates significant efficiency challenges. FlashAttention~\cite{dao2022flashattention,dao2023flashattention2,shah2024flashattention} addresses these limitations through algorithmic innovations that optimize memory access patterns and reduce unnecessary memory reads/writes during attention computation. By leveraging tiling strategies and fusing operations to maximize data reuse within fast GPU memory hierarchies, FlashAttention substantially improves throughput and enables processing of longer sequences without compromising model quality. These efficiency gains have been instrumental in scaling transformer models to increasingly larger contexts and higher resolution inputs, but the fundamental quadratic complexity of attention remains an inherent limitation.

\noindent\textbf{Sequence Modeling in 1D and 2D Space.} Sequential modeling has been dominated by recurrent architectures like LSTMs~\cite{hochreiter1997long}, GRUs~\cite{chung2014empirical}, and 2D-LSTMs~\cite{graves2007multi,byeon2015scene}, which process data through non-linear transformations. Despite their effectiveness, these approaches face fundamental limitations in computational efficiency and scalability due to their inherent sequential nature. Their non-linear propagation mechanisms also struggle with long-term dependencies, often suffering from gradient vanishing or exploding issues~\cite{hochreiter1991untersuchungen,pascanu2013difficulty} that prevent distant information from effectively influencing future states. State Space Models (SSMs) have emerged as promising alternatives to attention-based architectures, offering linear computational complexity with respect to sequence length. Pioneering approaches like S4~\cite{gu2021efficiently} and Mamba~\cite{gu2023mamba} implement continuous-time dynamical systems through discretized state spaces, enabling efficient modeling of long-range dependencies without the quadratic cost of attention. These models maintain a compact hidden state that evolves through linear recurrence relations, often employing selective scanning mechanisms to adapt to input-dependent patterns. For visual tasks, several approaches~\cite{nguyen2022s4nd,baron20232,zhu2024ViM,liu2024vmamba,li2024mamba} have adapted SSMs by linearizing 2D image data, though this transformation potentially compromises inherent spatial relationships present in the original data structure.

\noindent\textbf{Spatial Propagation Networks.} The Spatial Propagation Network (SPN)~\cite{liu2017learning} pioneered linear propagation specifically for 2D data, initially designed as a single-layer component on top of CNNs for sparse-to-dense prediction tasks like segmentation. However, SPN's potential as a scalable foundational architecture comparable to Vision Transformers (ViT) remains largely unexplored. Moreover, SPN's sequential processing across different spatial directions inherently limits its computational efficiency, and it fails to adequately address long-range propagation requirements crucial for high-level vision tasks. Our GSPN architecture advances beyond these limitations by implementing parallel row/column-wise propagation mechanisms that enable efficient learning of affinity matrices while maintaining gradient stability and effective long-range correlation. Through both theoretical analysis and empirical evaluation, we demonstrate that GSPN represents a compelling alternative to established ViT and Mamba architectures.

\section{Background}
\label{sec:background}

We introduce the background of the propagation algorithm itself and the GPU design principles. In~\Cref{sec:gpu}, we review modern A100 architecture—the grid/block/warp execution model, on-chip shared memory, and high-bandwidth device memory. We explain how these features shape kernel performance. In~\Cref{subsec:linear_prop}, we review GSPN's line-scan propagation formulation. The final section summarizes how this recurrence is mapped onto CUDA blocks in a custom kernel implemented in \cite{wang2025parallel} to realize parallelism.

\subsection{GPU Hardware Characteristics}\label{sec:gpu}
\label{subsec:hardware}

On modern NVIDIA GPUs like the A100, computation is dispatched as a grid of thread blocks. Each block can be 1D, 2D, or 3D (e.g., \texttt{blockDim.x} alone for 1D block, or both \texttt{blockDim.x} and \texttt{blockDim.y} for 2D block). 
Inside each block, threads are organized into warps of 32 threads each.
The total number of warps per block depends on the block’s thread count (for example, a block with 1024 threads has 32 warps).
To maximize throughput, blocks must be sized to supply enough active warps per Streaming Multiprocessors (SM) without exceeding their on-chip shared-memory or register limits--this balance is what drives high occupancy.
Within each block, threads share a small SRAM buffer ("shared memory") for low-latency reuse, while all other tensors are streamed in and out of off-chip high-bandwidth memory (HBM) through the L2/L1 caches.

\subsection{2D Spatial Propagation Algorithm Overview}
\label{subsec:linear_prop}

Generalized Spatial Propagation Networks (GSPN)~\cite{wang2025parallel} perform 2D spatial modeling through row-by-row (or column-by-column) linear propagation. For an input image $x \in \mathbb{R}^{H \times W \times C}$, this involves processing one dimension (e.g., rows) sequentially, while computations within each step (e.g., along a row) are parallelized.
Focusing on row-wise propagation, where $i\in [0, H-1]$ is the row index, let $h_{i,:,c} \in \mathbb{R}^W$ be the hidden state for row $i$ and channel $c$, $x_{i,:,c} \in \mathbb{R}^W$ be the corresponding input row, and $\lambda_{i,:,c} \in \mathbb{R}^W$ be a learnable, input-dependent scaling vector for channel $c$ at row $i$, 
and $w_{i,c} \in \mathbb{R}^{W \times W}$ be a learnable, per-channel propagation weight matrix for row $i$ and channel $c$. The per-channel recurrence is:
\begin{equation}
h_{i,:,c} = w_{i,c} h_{i-1,:,c} + \text{Diag}(\lambda_{i,:,c}) x_{i,:,c}
\label{eq:gspn_recurrence}
\end{equation}
The per-channel hidden state $h_{i,:,c}$ is then transformed by an output layer. Let $u_{i,:,c} \in \mathbb{R}^W$ be a learnable output vector. The final output for row $i$ and channel $c$ is given by:
\begin{equation}
y_{i,:,c} = u_{i,:,c} \odot h_{i,:,c}
\label{eq:gspn_output_y}
\end{equation}
Guided by the Stability-Context Condition introduced in~\cite{wang2025parallel}, $w_{i,c}$ is learned and normalized to be row-stochastic—every row sums to 1—thereby guaranteeing numerical stability while still capturing long-range context. In addition, each element in row $i$ connects to only three neighboring elements in the previous row $i-1$ (e.g., top-left, top-center, top-right for top-to-bottom propagation), resulting in $w_{i,c}$ being a tridiagonal matrix. A single pass of this recurrence connects pixels within a local region. To achieve full-grid propagation, GSPN performs four complementary directional passes—top-to-bottom, bottom-to-top, left-to-right, and right-to-left. By combining the 3-neighbor kernel with these four passes, the model attains dense pairwise connectivity across the image while remaining efficient, since only three coefficients are learned per pixel.

This row-wise propagation requires $O(H)$ sequential steps, where within each step, all $W$ elements of the current row are computed in parallel (or vice versa for column-wise propagation), yielding an $O(\max(H,W))$ computational complexity—equivalent to $O(\sqrt{N})$ for square images with $N$ pixels. GSPN also offers a local variant, i.e., GSPN-local, which splits each row or column into fixed-length segments of size 
\texttt{kchunk} and confines propagation to within those segments.
Finally, this recurrence is related to linear attention. Let $\Lambda_i = \text{blockdiag}(\text{Diag}(\lambda_{i,:,0}), \dots, \text{Diag}(\lambda_{i,:,C-1}))$. The overall operation is $y_i = u_i\sum_{j=0}^{i-1} (\prod_{\tau=j+1}^{i} w_\tau) \Lambda_j x_j$
which resembles a linear attention mechanism, with $\prod_{\tau=j+1}^{i} w_\tau$ the normalization term.

\begin{figure}[t]
    \centering
    \includegraphics[width=\textwidth]{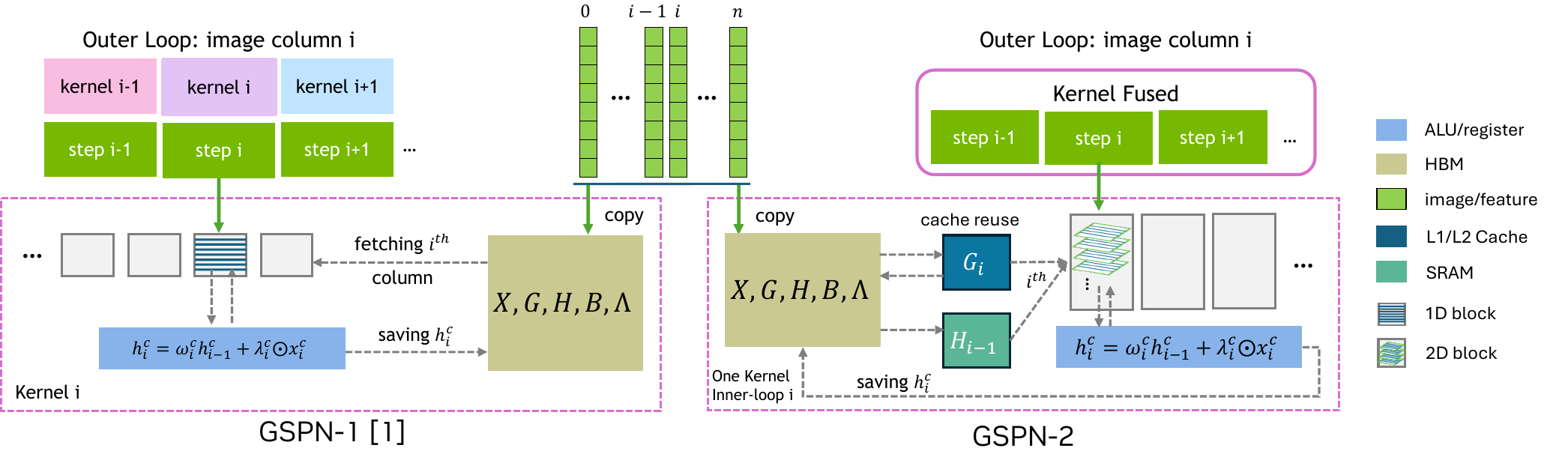}
    \caption{\textbf{Pipeline optimization from GSPN-1 to GSPN-2.} GSPN-1 launches separate kernels for each image column $i$, leading to redundant HBM access and limited temporal data reuse. Each kernel independently computes $h_i^c = \omega_i^c h_{i-1}^c + \lambda_i^c \odot x_i^c$, fetching and storing intermediate states via global memory. GSPN-2 fuses these operations into a single kernel with an inner loop over columns, enabling cache and register reuse of $h_{i-1}^c$, $G_i$, and other temporaries. This design minimizes memory traffic, maximizes locality, and leverages shared memory for efficient on-chip computation.}%
    \label{fig:cuda_kernel}
\end{figure}

\subsection{CUDA Implementation in GSPN \cite{wang2025parallel}}\label{sec:cuda_org}

The baseline GSPN implementation, referred to as GSPN-1, maps the 2D spatial propagation (Eq.~\ref{eq:gspn_recurrence}) to CUDA by iterating sequentially over the propagation dimension (e.g., height $H$) while parallelizing computations across the orthogonal dimension (e.g., width $W$). To handle the inherent sequential dependency along the propagation axis (e.g., rows $i=0, \dots, H-1$), it launches separate, relatively lightweight CUDA kernels for individual steps or small chunks, as illustrated in Figure~\ref{fig:cuda_kernel}(a).

For each step in the propagation sequence (e.g., processing a column based on the previous one in a left-to-right scan), GSPN-1 launches a new CUDA kernel, which introduces significant kernel-launch overhead along the propagation direction. Within each kernel, computations are parallelized across the orthogonal spatial dimension (width $W$), batches ($N$), and channels ($C$) by flattening these dimensions into a 1D grid of thread blocks (typically \texttt{blockDim.x = 512}). However, this simplistic mapping ignores CUDA’s warp-level scheduling and often results in suboptimal hardware utilization. Moreover, all tensors—including inputs ($x$), hidden states from the previous step ($h_{i-1,:,c}$), learnable weights ($w_{i,c}, \lambda_{i,:,c}$), and the current outputs ($h_{i,:,c}$)—are repeatedly read from and written back to global GPU memory (HBM), causing high latency and minimal data reuse in on-chip memory. These issues—frequent kernel launches, suboptimal thread mapping, and excessive global-memory access—together limit the efficiency of GSPN-1, motivating the redesign presented in Section~\ref{sec:gspn2_method}.

\section{GSPN-2: Efficient Algorithm and System Co-design} 
\label{sec:gspn2_method}
While the baseline GSPN-1 implementation established functional correctness, its CUDA design suffered from inefficiencies—frequent kernel launches, flat 1D block configurations, and unpredictable memory reuse, which led to suboptimal data locality and high launch overhead. To address these limitations, we redesign both the algorithm and its GPU execution pipeline through GSPN-2, focusing on three principles: (1) a single-kernel propagation scheme that eliminates redundant launches, (2) channel-compressive propagation with shared weights and proxy compression to reduce concurrency load, and (3) optimized CUDA execution leveraging shared memory, coalesced access, and stream-level parallelism.
This section introduces the evolution of GSPN-2, from its single-kernel design to memory- and concurrency-aware optimization.

\subsection{A Single-Kernel Design}
\label{subsec:single_kernel}

\begin{wrapfigure}{r}{0.5\textwidth}
  \centering
  \includegraphics[width=\linewidth]{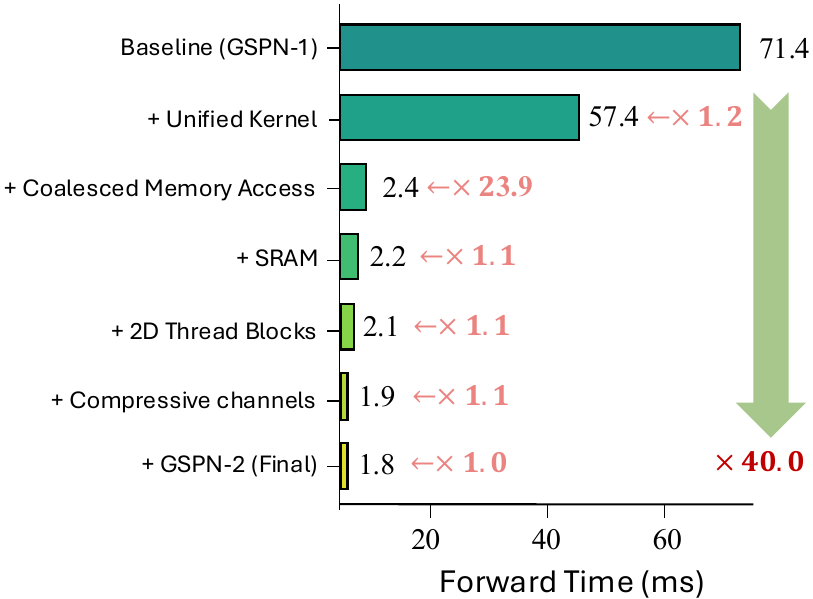}
  \caption{\textbf{Step-by-step optimization of the GSPN CUDA kernel.} Each bar shows the reduction in forward time (ms) achieved through cumulative optimizations, starting from the GSPN-1 baseline. The final implementation (GSPN-2) achieves a 40.0× speedup compared to the baseline.}
  \label{fig:journey}\vspace{-5mm}
\end{wrapfigure}

\paragraph{Kernel Fuse.} 
We consolidate these numerous small kernels into a single, unified CUDA kernel. This single kernel is designed to process the entire outer-loop (e.g., all columns in a left-to-right scan) \textit{within} the kernel, while still parallelizing computations across the other dimensions (batch, channels, and rows/height). By eliminating thousands of micro-launches, this single-kernel approach drastically reduces launch overhead. 
For instance, preliminary tests showed that simply moving from a multi-kernel to a single-kernel implementation for a typical GSPN configuration immediately yielded a significant performance boost (e.g. 1.2× faster) in~\Cref{fig:journey}, even before other memory or algorithmic optimizations were applied.
We illustrate substantial performance gains from this and subsequent optimization stages across various hardware configurations and input dimensions (batch size, channels, height, width) in Figure~\ref{fig:journey} and Section~\ref{subsec:profiling}.

\paragraph{Block and Grid Configuration.} 
In GSPN-1, the kernel used a flat 1D grid of blocks (\texttt{blockDim.x = 512}) where threads were linearly mapped across combinations of batch ($N$), channel ($C$), height ($H$), and chunk index ($k_{\text{chunk}}$). This configuration resulted in insufficient locality and suboptimal warp utilization. 
In GSPN-2, the CUDA grid is indexed by the tuple $(\text{chunk}, n, c)$, so that each block corresponds to one unique $(\text{chunk}, n, c)$ combination and processes a full spatial column along height. The grid therefore contains $k_{\text{chunk}} \times N \times C$ blocks in total, which can be realized as a 1D grid or a 3D grid to respect CUDA’s per-axis limits. Each block uses up to 1024 threads along the height dimension. For $H \le 1024$, one thread is assigned per row, achieving full occupancy. When $H > 1024$, threads iterate over multiple rows with stride \texttt{blockDim.x}, ensuring complete coverage without exceeding the per-block thread limit.

\subsection{Compact Channel Propagation}
\label{sec:gspnv2_theory}
A key performance bottleneck in GSPN-1 arises from GPU concurrency saturation when the number of active CUDA blocks—proportional to $k_{\text{chunk}} \times N \times C$—exceeds the hardware's concurrent execution capacity. On GPUs such as NVIDIA A100, each Streaming Multiprocessor (SM) can host up to 32 resident thread blocks (compute capability 8.0), and with 108 SMs available, roughly $108 \times 32 \approx 3{,}500$ blocks can be active concurrently under ideal conditions. 
Under typical GSPN workloads, kernel execution time remains nearly constant up to this scale (about 3--4K concurrent blocks). Beyond that point, the runtime grows linearly as additional blocks wait in the scheduling queue. This saturation effect causes GSPN-1 to lose its near-constant runtime scaling when operating on high-dimensional feature maps (e.g., thousands of channels).

To address this, GSPN-2 introduces a \textbf{compact multi-channel propagation} scheme that reduces the effective channel concurrency while maintaining expressive multi-channel behavior. The core idea is to project the input tensor $x \in \mathbb{R}^{N \times C \times H \times W}$ into a lower-dimensional proxy subspace $x_{\mathrm{proxy}} \in \mathbb{R}^{N \times C_{\mathrm{proxy}} \times H \times W}$, where $C_{\mathrm{proxy}} \ll C$ (e.g., $C_{\mathrm{proxy}} = 8$). The propagation is then applied to this proxy representation using shared propagation matrices $w_i$ and later restored to the original $C$-channel space. This reduces the total block count from $k_{\text{chunk}} \times N \times C$ to $k_{\text{chunk}} \times N \times C_{\mathrm{proxy}}$, reducing it as much as possible to stay well within the hardware concurrency regime (roughly 3--4K on A100-class GPUs) and thereby sustaining near-constant performance.

\paragraph{Illustrative Single-Channel Case.} 
We use the single-channel case to make the attention analogy explicit. We replace per-channel weights with a single propagation matrix per column. In this view, $w_i$ will be shared among all the channels, which plays the role of an attention-style affinity matrix over positions in column $i$, and the per-position input scaling acts like value gating. The per-channel recurrence thus becomes:
\begin{equation}
h_{i,:,c} = w_i h_{i-1,:,c} + \lambda_{i,:,c} \odot x_{i,:,c} = w_i h_{i-1,:,c} + \text{Diag}(\lambda_{i,:,c}) x_{i,:,c}
\end{equation}\label{eq:2dsp}
where $w_i$ governs spatial propagation along the column, and $\lambda_{i,:,c}$ preserves per-channel modulation. 
This formulation significantly reduces the number of parameters while retaining the same functional structure. Stacking all channels, the full recurrence $h_i = W_i h_{i-1} + \Lambda_i x_i$ still holds, now with channel-shared $w_i$. 
To expand Eq. \eqref{eq:2dsp}, we denote $H_v, X_v$ as the concatenation of all $h_{i,:,c}$ and $x_{i,:,c}$ into vectors. The expansion yields a block lower-triangular matrix form:
\begin{small}
	\begin{equation}
	H_v =
	\left[
	\begin{matrix}
	\Lambda_1     & 0    & \cdots & \cdots & 0    \\
	w_2\Lambda_1    & \Lambda_2  & 0 & \cdots & 0   \\
	w_3w_2\Lambda_1     & w_3\Lambda_2    & \Lambda_3  &  0  & \cdots 		\\
	\vdots &   \vdots   & \vdots & \ddots & \vdots    \\
	(\prod_{k=2}^{L} w_k)\Lambda_1 & (\prod_{k=3}^{L} w_k)\Lambda_2 & \cdots & w_L\Lambda_{L-1} & \Lambda_L    \\
	\end{matrix}
	\right]X_v	\\
	=GX_v,
	\label{eq:global}
	\end{equation}
\end{small}
where each block $G_{ij}$ is a $N \times N$ matrix representing how the input slice $x_j$ contributes to the output $h_i$—directly analogous to an attention mechanism’s affinity matrix.
Here, the channel-shared matrices $w_i$ define dense spatial relationships, while the channel-specific scaling matrices $\Lambda_j$ inject feature-wise modulation. This formulation shows that, in the single-channel case, GSPN-2 can be viewed as an attention-like process with learnable spatial affinities.

\paragraph{Compressive Proxy Dimension.}
To further relieve concurrency saturation when $N\times C$ is large, we compress the channel axis before propagation. Concretely, we project $x\in\mathbb{R}^{N\times C\times H\times W}$ to $x_{\mathrm{proxy}}\in\mathbb{R}^{N\times C_{\mathrm{proxy}}\times H\times W}$ with $C_{\mathrm{proxy}}\ll C$ (e.g., $C_{\mathrm{proxy}}{=}8$), apply the same columnwise recurrence in the proxy space using the shared $w_i$, and expand back to $C$ with a learned $1\times1$ projection. This reduces the grid from $k_{\text{chunk}}\times N\times C$ to $k_{\text{chunk}}\times N\times C_{\mathrm{proxy}}$, shrinking the number of simultaneously scheduled block slices (e.g., $N\times C_{\mathrm{proxy}}\times H$ for a row scan). We choose $C_{\mathrm{proxy}}$ to minimize the active-block budget and delay entry into the post-saturation, near-linear regime on A100-class GPUs; even when that plateau cannot be fully avoided (very large $N$), the compression still cuts queueing and improves SM utilization while preserving multi-channel expressiveness.

\subsection{Efficient CUDA Scaling under Large Block-Slice Loads}
\label{subsec:gspn2_cuda_opts}
This section presents CUDA kernel enhancements—particularly grid/block reconfiguration and on-chip memory strategies—that enable efficient compact channel propagation even when the block count ($k\_{\text{chunk}}\times N\times C$) becomes very large.

\paragraph{\textbf{SRAM for Hidden States.}}
We cache the previous step’s hidden state (e.g., $h_{i-1}$) in on-chip shared memory to reduce redundant global-memory (HBM) reads. Within each CUDA block, threads cooperatively process a \emph{tile}—a small subset of spatial positions or channel slices—and reuse the cached hidden-state values stored in shared memory. This on-chip reuse reduces latency when multiple threads within a block access overlapping regions of $h_{i-1}$, such as along a spatial column. The performance gain depends on configuration: it is most effective when the reuse per tile is high, the shared-memory footprint fits comfortably within per-block limits, and bank conflicts are minimal. When reuse is low or L1/L2 caching already covers the working set, the benefit diminishes. Therefore, we enable this shared-memory caching selectively and tune tile size and \texttt{cSlice} to balance reuse against occupancy.

\paragraph{\textbf{2D Block Design for Channel-Parallel Propagation.}}
Building upon the 1D block design in Sec.~\ref{subsec:single_kernel}, we extend it to a 2D configuration by introducing a second dimension, \texttt{cSlice}, such that each CUDA block has $blockDim = (H, cSlice)$. Within a block, \texttt{threadIdx.x} corresponds to spatial positions along a column (up to $H$), while \texttt{threadIdx.y} spans a small group of channel slices. This enables the block to process multiple channels of the same column in parallel, improving hardware utilization and memory throughput even when each channel maintains its own propagation weight $w_i^c$. Compared to the earlier 1D block layout, this 2D configuration achieves better occupancy and reduced latency by aligning computation and memory access patterns across both spatial and channel dimensions, as demonstrated in~\Cref{subsec:profiling}.

\paragraph{\textbf{Coalesced Memory Access.}}
A major source of speedup in GSPN-2 comes from enforcing coalesced global-memory access. We arrange $x_i$, $h_i$, and $w_i$ contiguously in memory so that consecutive threads in a warp access adjacent addresses when reading or writing. This pattern allows the hardware to combine multiple per-thread transactions into a single wide memory operation, fully utilizing the available bandwidth and minimizing wasted cycles. By eliminating the irregular, strided accesses present in GSPN-1, the coalesced layout contributes the largest single performance gain among all CUDA-level optimizations (see Figure~\ref{fig:journey}).

\paragraph{\textbf{Stream-Based Concurrency.}}
For multi-directional propagation, GSPN-2 executes each directional pass on a separate, non-blocking CUDA stream. This allows concurrent kernel execution across directions, improving hardware utilization by keeping more SMs active—especially on GPUs with abundant SM resources. The benefit depends on workload balance and available parallelism; it is most pronounced when the directional passes have similar compute and memory footprints. In addition, when the grid dimension exceeds CUDA’s per-axis limit of 65,535, GSPN-2 automatically performs multiple launches with offset indexing to cover the excess range without interrupting the overall stream concurrency.

\section{Experiments}
To demonstrate the effectiveness of GSPN-2, we design experiments that answer two questions: \textit{How much faster is it?} and \textit{Does the speed-up preserve—or even improve—task performance?} We first profile the new CUDA kernel, isolating the gains from different factors, such as unified kernel launch, shared memory, and channel-share weights. We then benchmark GSPN-2 on a suite of vision tasks, comparing accuracy and throughput against GSPN-1 and other strong baselines. By evaluating both efficiency and task performance, we demonstrate the benefits of our tightly integrated algorithm and kernel optimizations.

\subsection{Detailed Profiling and Performance Characteristics}
\label{subsec:profiling}

To understand GSPN-2's performance characteristics in depth, we conducted comprehensive profiling across various input configurations, analyzing memory throughput, cache utilization, and computational efficiency.

\paragraph{Step-by-step CUDA Optimization.} We benchmark a typical configuration, i.e., 1024$\times$1024 image size, batch size 16, 8 channels, and quantify the impact of each CUDA kernel optimization term in Figure~\ref{fig:journey}. The GSPN-1 baseline exhibited suboptimal performance (71.4 ms) due to kernel launch overhead and inefficient memory access patterns. Our first optimization—a single fused kernel (Sec.~\ref{subsec:single_kernel})—eliminates thousands of micro-launches by processing entire scan operations within a single kernel, yielding a notable 1.2$\times$ speedup (57.4 ms). \textbf{Coalesced Memory Access} patterns (Sec.~\ref{subsec:gspn2_cuda_opts}) maximized memory bandwidth utilization for a substantial 23.9$\times$ improvement (2.4 ms). Implementing \textbf{Shared Memory Cache} for hidden states (Sec.~\ref{subsec:gspn2_cuda_opts}) reduced global memory traffic by 1.1$\times$ (2.2 ms). Restructuring to \textbf{2D Thread Blocks} (Sec.~\ref{subsec:single_kernel}, Sec.~\ref{subsec:gspn2_cuda_opts}) improved thread organization and data locality for another 1.1$\times$ gain (2.1 ms). \textbf{Compressive channels} (Sec .~\ref{sec:gspnv2_theory}) reduced parameter fetch overhead and enhanced cache coherence for a 1.1$\times$ speedup (1.9 ms). The fully optimized GSPN-2 implementation achieves an impressive 40.0$\times$ cumulative speedup (1.8 ms) over the original baseline. We note that the relative impact of each optimization varies with workload characteristics (batch size, channel count); Section~\ref{app:effablat_large_batch} in the appendix provides a detailed analysis under an alternative large-batch configuration (batch size 256, 1 channel), demonstrating that while coalesced memory access remains the dominant optimization, \textbf{Shared memory caching} and \textbf{2D thread blocks} (Sec.~\ref{subsec:gspn2_cuda_opts}) exhibit configuration-dependent benefits.

\paragraph{\textbf{Memory Throughput Analysis.}}
As shown in~\Cref{fig:kernel_throughput}, NVIDIA Nsight Compute profiling indicates that GSPN-2 achieves memory throughput near the theoretical limit, with global-memory efficiency reaching 93\% on A100 GPUs. This efficiency remains remarkably stable across a wide range of batch sizes and spatial resolutions, demonstrating effective saturation of the available bandwidth. In contrast, GSPN-1 exhibits highly variable throughput—only $3$--$8\%$ of peak—that further deteriorates as input dimensions increase.

\begin{table}[h]
\rowcolors{2}{ngreen!15}{white}
\caption{\textbf{Global memory throughput under typical input configurations on A100 GPU.} We show throughput for a range of input sizes, batch sizes, and channel counts representative of common deployment scenarios in different tasks. Rather than exhaustively sweeping all variables, we select practical configurations to demonstrate consistent and significant gains of GSPN-2 over GSPN-1 across diverse settings.} 

\centering
\begin{tabular}{c|c|c|c|c}
\toprule
\rowcolor{ngreen!40}
\textbf{Input Size} & \textbf{Batch} & \textbf{Channels} & \textbf{GSPN-1 Throughput} & \textbf{GSPN-2 Throughput} \\
\midrule
32×32 & 32 & 196 & 114 GB/s (6.0\%) & 1832 GB/s (91.8\%) \\
64×64 & 1 & 768 & 86 GB/s (4.5\%) & 1847 GB/s (92.3\%) \\
64×64 & 1 & 1152 & 35 GB/s (2.1\%) & 1837 GB/s (92.0\%) \\
64×64 & 1 & 32 & 125 GB/s (6.3\%) & 1830 GB/s (91.5\%) \\
128×128 & 1 & 32 & 98 GB/s (4.9\%) & 1865 GB/s (93.3\%) \\
256×256 & 1 & 64 & 76 GB/s (3.8\%) & 1842 GB/s (92.1\%) \\
256×256 & 8 & 64 & 94 GB/s (4.7\%) & 1858 GB/s (92.9\%) \\
512×512 & 1 & 128 & 64 GB/s (3.2\%) & 1840 GB/s (92.0\%) \\
\bottomrule
\end{tabular}
\label{fig:kernel_throughput}
\end{table}

\begin{figure}[t]
    \centering
    \includegraphics[width=\textwidth]{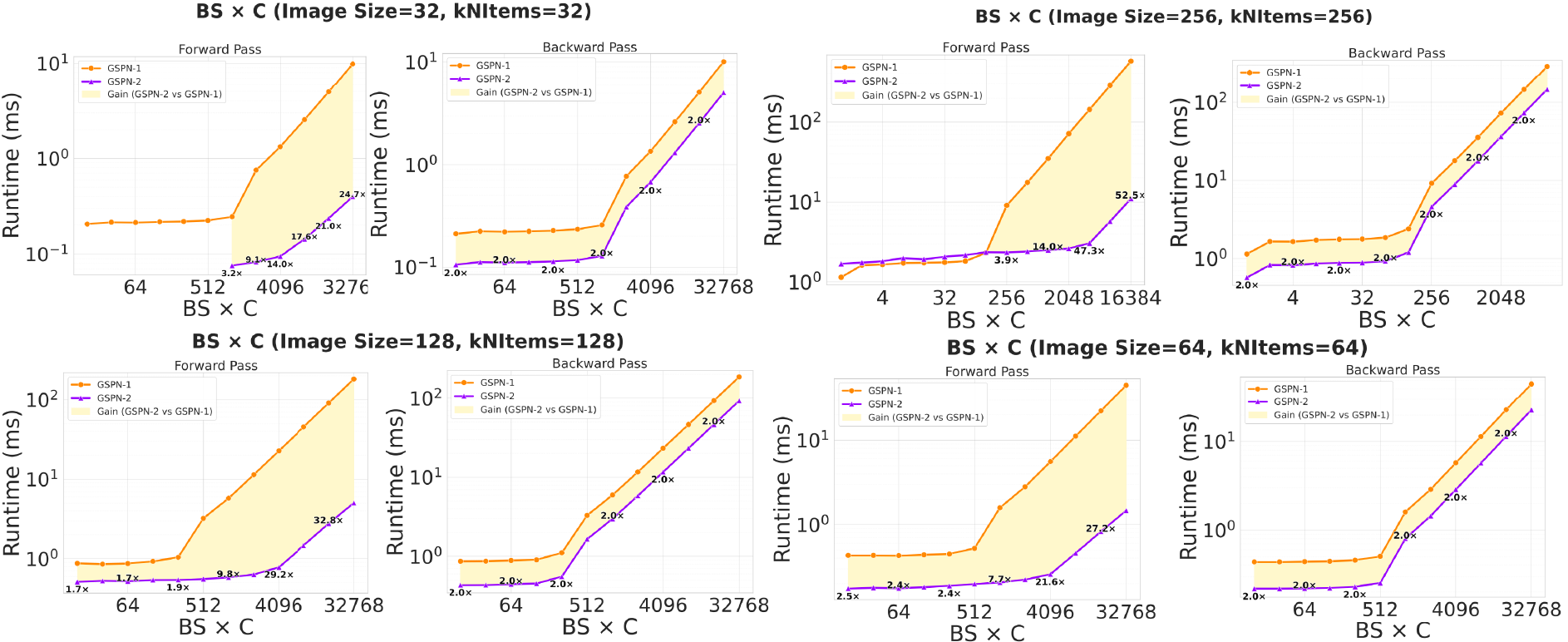}
    \caption{\textbf{Runtime Performance Comparison of GSPN-1 and GSPN-2.} We show forward and backward pass execution times (in milliseconds) across different channel counts. Results are presented for various configurations. GSPN-2 greatly improves the runtime of both forward and backward passes across different cases.} %
    \label{fig:runtime}
\end{figure}

\paragraph{Performance Scaling with Input Size.} As shown in the upper row of Figure~\ref{fig:runtime}, GSPN-2 consistently outperforms GSPN-1 across various image resolutions with fixed batch and channel counts. For large image sizes (1024×1024), we observe speedups of up to 36.8× for forward passes and 25.3× for backward passes. This substantial improvement is particularly relevant for high-resolution visual processing tasks such as image generation and super-resolution, where spatial dimensions significantly impact computational demands. The performance gap widens as image resolution increases, highlighting GSPN-2's superior ability to handle spatially dense computations efficiently through its optimized memory access patterns and unified kernel design.

\paragraph{Performance with Varying Batch Size and Channel Dimensions.} The lower row of Figure~\ref{fig:runtime} demonstrates GSPN-2's exceptional performance in scenarios requiring large batch sizes or high channel dimensions—critical requirements for video generation, foundation model visual towers, and multimodal applications. With three distinct performance lines (GSPN-1 and GSPN-2), we observe that GSPN-2 maintains consistent 2-4× speedups over GSPN-1 even as batch sizes scale to 256 or channel counts increase to 1024. For instance, when processing inputs with 256 channels, GSPN-2 achieves a 27.4× speedup on forward passes and 48.6× on backward passes. These improvements are particularly valuable for production inference systems handling multiple streams simultaneously or for models requiring high feature dimensionality. The channel-sharing approach (Sec.~\ref{sec:gspnv2_theory}) provides additional efficiency gains of up to 1.5× in these demanding scenarios, enabling practical deployment of GSPN architectures in compute-intensive applications like real-time video processing and multimodal foundation models.

\paragraph{L1 Cache Effectiveness.}
One surprising finding from our profiling is the effectiveness of the L1 cache even without explicit shared memory caching in certain configurations. When we experimented with a shared memory buffer to store previous hidden states ($h_{t-1}$), we observed that performance remained largely unchanged compared to relying on L1 cache. Detailed profiling revealed L1 cache hit rates of approximately 35\% for the standard implementation. Interestingly, when using shared memory explicitly, L1 hit rates dropped to near 0\%, with those accesses now served from shared memory instead. Despite this shift in memory hierarchy usage, latency remained comparable between both approaches, suggesting modern GPU L1 caches are highly effective for structured access patterns. The transposed data layout and coalesced access patterns enable effective hardware caching, even without explicit shared memory management in some cases. However, for maximum portability across GPU architectures and to ensure deterministic performance, the shared memory implementation remains preferable.

\paragraph{Streaming Multiprocessor Utilization}
Our profiling reveals an important relationship between input configuration and SM utilization. With GSPN-2's 2D thread organization strategy, SM occupancy varies significantly based on workload characteristics. For large batch sizes and channel counts, SM occupancy approaches 100\%, fully utilizing the GPU's 108 SMs on A100. However, for small batch sizes and channel counts, occupancy can drop significantly (as low as 20-30\%). This occurs because when processing independent chunks, each chunk requires one block, limiting parallelism for small input dimensions. This suggests potential areas for further optimization—particularly for low batch size, low channel count scenarios where we could further decompose the problem to increase parallelism across SMs.

\begin{table*}[h!]
\renewcommand{\arraystretch}{1.5}
    \caption{
    \textbf{Performance of models on ImageNet at the resolution of $224^2$.} Colors denote different backbone types: \textbf{\textcolor{yellow}{yellow}} for CNNs, \textbf{\textcolor{RPurple}{orange}} for Transformers, and \textbf{\textcolor{Green}{green}} for Raster scan (i.e., 1D linear propagation) methods. %
    }
    \label{tab:cls-imagenet}
    \centering
    \begin{minipage}{.3\textwidth}
    \tiny
    \centering
    \setlength{\tabcolsep}{.1pt}
    \begin{tabular}{l | c | c | c c }
    \toprule
    \multirow{3}{*}{\makecell[c]{Model}}    & \multirow{3}{*}{\makecell[c]{Backbone}}     & \multirow{3}{*}{\makecell[c]{Param \\ (M)}}   & \multicolumn{2}{c}{IN-1K} \\
    \cline{4-5}
    ~ & ~ & ~ &  \multirow{2}{*}{\makecell[c]{MAC \\ (G)}} & \multirow{2}{*}{\makecell[c]{Acc \\ (\%)}} \\ 
    ~ & ~ & ~ & ~ & ~ \\
    \midrule
    \rowcolor{Yellow}
    ConvNeXT-T \cite{liu2022convnet} &  CN   & 29 & 4.5 & 82.1  \\
    \rowcolor{Yellow}
    MambaOut-Tiny \cite{yu2024mambaout} & CN & 27 & 4.5 & 82.7 \\
    \rowcolor{RPurple}
    DeiT-S \cite{touvron2021training} &  TF  & 22 & 4.6 & 79.8  \\
    \rowcolor{RPurple}
    T2T-ViT-14 \cite{yuan2021tokens} &  TF   & 22 & 4.8 & 81.5 \\
    \rowcolor{RPurple}
    Swin-T \cite{liu2021swin} &  TF   & 29 & 4.5 & 81.3 \\
    \rowcolor{RPurple}
    SwinV2-T \cite{liu2021swinv2} & TF & 28 & 4.4 & 81.8 \\
    \rowcolor{RPurple}
    CSWin-T \cite{dong2022cswin} &  TF  & 23 & 4.3 & 82.7  \\
    \rowcolor{RPurple}
    CoAtNet-0 \cite{dai2021coatnet} &   TF  & 25 & 4.2 & 81.6 \\
    \rowcolor{Green}
    Vim-S \cite{zhu2024ViM} & RS & 26 & 5.1 & 80.5 \\
    \rowcolor{Green}
    VMamba-T \cite{liu2024vmamba} &  RS & 22 & 5.6 & 82.2 \\
    \rowcolor{Green}
    Mamba-2D-S \cite{li2024mamba} &  RS & 24 & -- & 81.7 \\
    \rowcolor{Green}
    LocalVMamba-T \cite{huang2024localmamba} & RS & 26 & 5.7 & 82.7 \\
    \rowcolor{Green}
    VRWKV-S \cite{duan2024visionrwkv} & RS & 24 & 4.6 & 80.1 \\
    \rowcolor{Green}
    ViL-S \cite{alkin2024visionlstm} & RS & 23 & 5.1 & 81.5 \\
    \rowcolor{Green}
    MambaVision-T \cite{hatamizadeh2024mambavision} & RS & 32 & 4.4 & 82.3 \\
    \midrule
    \rowcolor{Blue}
    GSPN-T & Line & 30 & 5.3 & \textbf{83.0} \\
    \rowcolor{Blue}
    \textbf{GSPN-2-T (Ours)} & Line & 24 & 4.2 & \textbf{83.0} \\
    \bottomrule
    \end{tabular}
    \end{minipage}
\renewcommand{\arraystretch}{1.3}
    \begin{minipage}{.3\textwidth}
    \tiny
    \centering
    \setlength{\tabcolsep}{.1pt}
    \begin{tabular}{l | c | c | c c }
    \toprule
    \multirow{3}{*}{\makecell[c]{Model}}    & \multirow{3}{*}{\makecell[c]{Backbone}}     & \multirow{3}{*}{\makecell[c]{Param \\ (M)}}   & \multicolumn{2}{c}{IN-1K} \\
    \cline{4-5}
    ~ & ~ & ~ &  \multirow{2}{*}{\makecell[c]{MAC \\ (G)}} & \multirow{2}{*}{\makecell[c]{Acc \\ (\%)}} \\ 
    ~ & ~ & ~ & ~ & ~ \\
    \midrule
    \rowcolor{Yellow}
    ConvNeXT-S \cite{liu2022convnet} & CN & 50 & 8.7 & 83.1 \\
    \rowcolor{Yellow}
    CNFormer-S36 \cite{yu2024metaformer} & CN & 40 & 7.6 & 84.1 \\
    \rowcolor{Yellow}
    MogaNet-B \cite{li2024MogaNet} & CN & 44 & 9.9 & 84.3 \\
    \rowcolor{Yellow}
    InternImage-S \cite{wang2023internimage} & CN & 50 & 8.0 & 84.2 \\
    \rowcolor{Yellow}
    MambaOut-Small \cite{yu2024mambaout} & CN & 48 & 9.0 & 84.1 \\
    \rowcolor{RPurple}
    T2T-ViT-19 \cite{yuan2021tokens} & TF & 39 & 8.5 & 81.9 \\
    \rowcolor{RPurple}
    Focal-Small \cite{yang2022focal} & TF & 51 & 9.1 & 83.5 \\
    \rowcolor{RPurple}
    BiFormer-B \cite{zhu2023biformer} & TF & 57 & 9.8 & 84.3 \\
    \rowcolor{RPurple}
    NextViT-B \cite{li2022next} & TF & 45 & 8.3 & 83.2 \\
    \rowcolor{RPurple}
    Twins-B \cite{chu2021twins} & TF & 56 & 8.3 & 83.1 \\
    \rowcolor{RPurple}
    MaxViT-Small \cite{tu2022maxvit} & TF & 69 & 11.7 & \textbf{84.4} \\
    \rowcolor{RPurple}
    Swin-S \cite{liu2021swin} & TF & 50 & 8.7 & 83.0 \\
    \rowcolor{RPurple}
    SwinV2-S \cite{liu2021swinv2} & TF &  50 & 8.5 & 83.8 \\
    \rowcolor{RPurple}
    CoAtNet-1 \cite{dai2021coatnet} &  TF & 42 & 8.4 & 83.3 \\
    \rowcolor{RPurple}
    UniFormer-B \cite{li2022uniformer} &  TF & 50 & 8.3 & 83.9 \\
    \rowcolor{Green}
    VMamba-S \cite{liu2024vmamba} &  RS & 44 & 11.2 & 83.5 \\
    \rowcolor{Green}
    LocalVMamba-S \cite{huang2024localmamba} &  RS & 50 & 11.4 & 83.7 \\
    \rowcolor{Green}
    MambaVision-S \cite{hatamizadeh2024mambavision} & RS & 50 & 7.5 & 83.3 \\
    \midrule
    \rowcolor{Blue}
    GSPN-S & Line & 50 & 9.0 & 83.8 \\
    \rowcolor{Blue}
    \textbf{GSPN-2-S (Ours)} & Line & 50 & 9.2 & \textbf{84.4} \\
    \bottomrule
    \end{tabular}
    \end{minipage}
\renewcommand{\arraystretch}{1.35}
    \begin{minipage}{.3\textwidth}
    \tiny
    \centering
    \setlength{\tabcolsep}{.1pt}
    \begin{tabular}{l | c | c | c c }
    \toprule
    \multirow{3}{*}{\makecell[c]{Model}}    & \multirow{3}{*}{\makecell[c]{Backbone}}     & \multirow{3}{*}{\makecell[c]{Param \\ (M)}}   & \multicolumn{2}{c}{IN-1K} \\
    \cline{4-5}
    ~ & ~ & ~ &  \multirow{2}{*}{\makecell[c]{MAC \\ (G)}} & \multirow{2}{*}{\makecell[c]{Acc \\ (\%)}} \\ 
    ~ & ~ & ~ & ~ & ~ \\
    \midrule
    \rowcolor{Yellow}
    ConvNeXT-B \cite{liu2022convnet} & CN & 89 & 15.4 & 83.8 \\
    \rowcolor{Yellow}
    CNFormer-M36 \cite{yu2024metaformer} & CN & 57 & 12.8 & 84.5 \\
    \rowcolor{Yellow}
    MambaOut-Base \cite{yu2024mambaout} & CN & 85 & 15.8 & 84.2 \\
    \rowcolor{Yellow}
    SLaK-B\cite{liu2023more} & CN & 95 & 17.1 & 84.0 \\
    \rowcolor{RPurple}
    DeiT-B \cite{touvron2021training} & TF & 86 & 17.5 & 81.8 \\
    \rowcolor{RPurple}
    T2T-ViT-24 \cite{yuan2021tokens} & TF & 64 & 13.8 & 82.3 \\
    \rowcolor{RPurple}
    Swin-B \cite{liu2021swin} & TF & 88 & 15.4 & 83.5 \\
    \rowcolor{RPurple}
    SwinV2-B \cite{liu2021swinv2} & TF & 88 & 15.1 & \textbf{84.6} \\
    \rowcolor{RPurple}
    CSwin-B \cite{dong2022cswin} & TF & 78 & 15.0 & 84.2 \\
    \rowcolor{RPurple}
    MViTv2-B \cite{li2021improved} & TF & 52 & 10.2 & 84.4 \\
    \rowcolor{RPurple}
    CoAtNet-2 \cite{dai2021coatnet} &  TF & 75 & 15.7 & 84.1 \\
    \rowcolor{Green}
    Vim-B \cite{zhu2024ViM} & RS & 98 & 17.5 & 81.9 \\
    \rowcolor{Green}
    VMamba-B \cite{liu2024vmamba} &  RS & 89 & 15.4 & 83.9 \\
    \rowcolor{Green}
    Mamba-2D-B \cite{li2024mamba} &  RS & 92 & -- & 83.0 \\
    \rowcolor{Green}
    VRWKV-B \cite{duan2024visionrwkv} & RS & 94 & 18.2 & 82.0 \\
    \rowcolor{Green}
    ViL-B \cite{alkin2024visionlstm} & RS & 89 & 18.6 & 82.4 \\
    \rowcolor{Green}
    MambaVision-B \cite{hatamizadeh2024mambavision} & RS & 98 & 15.0 & 84.2 \\
    \midrule
    \rowcolor{Blue}
    GSPN-B & Line & 89 & 15.9 & 84.3 \\
    \rowcolor{Blue}
    \textbf{GSPN-2-B (Ours)} & Line & 89 & 14.2 & \textbf{84.9} \\
    \bottomrule
    \end{tabular}
    \end{minipage}
\end{table*}

\subsection{Image Classification}

In~\Cref{tab:cls-imagenet}, we present a comparative analysis of ImageNet-1K classification performance across three architectural paradigms: ConvNet-based~\cite{liu2022convnet,yu2024mambaout}, Transformer-based~\cite{touvron2021training,liu2021swin,dai2021coatnet,dong2022cswin,li2022next,li2022uniformer}, and sequential-based (RS scan) models~\cite{zhu2024ViM, liu2024vmamba,huang2024localmamba,li2024mamba,hatamizadeh2024mambavision, duan2024visionrwkv,alkin2024visionlstm} of varying sizes. %
For GSPN-2 models, the ImageNet experiments incorporate several key design choices: propagation weights $w_i$ are shared across channels in all GSPN modules, and a compressive proxy dimension $C_{\text{proxy}}$ is set to 2. This reduction in channel dimensionality allows the saved parameters to be reallocated for deeper or wider network architectures. Additionally, we integrate the Local Perception Unit (LPU)~\cite{guo2021cmt} at the beginning of each block and FFN. The MESA~\cite{du2022sharpness} technique is also employed to mitigate overfitting, contributing a further 0.2\% accuracy improvement to some variants.

Our GSPN-2 models, benefiting from the joint algorithmic and system-level redesign detailed in Section~\ref{sec:gspn2_method}, demonstrate notable advancements. The GSPN-2 series builds upon the strong foundation of GSPN-1, introducing refinements that enhance both performance and efficiency. GSPN-2-T achieves a competitive 83.0\% accuracy with significantly fewer parameters (24M vs. 30M for GSPN-T) and lower computational cost (4.2G MACs vs. 5.3G MACs for GSPN-T). It outperforms SSMs such as Vim-S (80.5\%), VMamba-T (82.2\%), and notably surpasses LocalVMamba-T (82.7\%) by 0.3\% accuracy with fewer MACs (4.2G vs 5.7G), while remaining competitive with leading ConvNets and Transformers in its category. GSPN-2-S achieves an impressive 84.4\% accuracy, marking a significant +0.6\% improvement over GSPN-S (83.8\%) with only a marginal increase in MACs (9.2G vs 9.0G) while using the same number of parameters (50M). This performance places GSPN-2-S ahead of strong competitors like MambaOut-Small (84.1\%) and UniFormer-B (83.9\%), showcasing its enhanced efficiency and effectiveness. At the base model scale, GSPN-2-B also achieves an excellent 84.9\% accuracy, improving upon GSPN-B (84.3\%) by +0.3\% while reducing MACs (14.2G vs 15.9G) with the same 89M parameters.

\subsection{Text-to-Image Generation}
\begin{figure}[t]
    \centering
    \includegraphics[width=\textwidth]{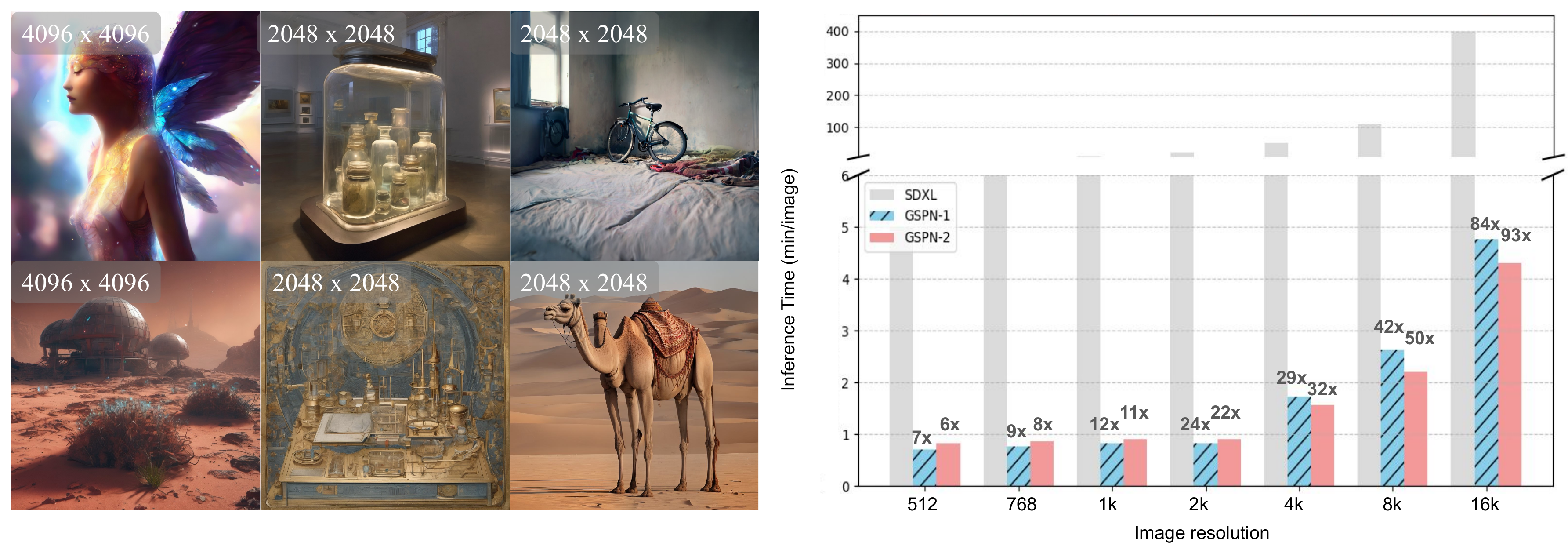}
    \caption{\textbf{Qualitative text-to-image results generated from our GSPN-2 SDXL model.} We enable generation up to 16K resolution on a single A100 GPU while reducing inference time by up to 93× on the SDXL model.} 
    \label{fig:vis1}
\end{figure}

To evaluate the efficiency and performance of GSPN-2 in high-resolution generative tasks, we conduct experiments on text-to-image generation using the Stable Diffusion XL (SDXL) framework, with results summarized in~\Cref{fig:vis1}.

Building upon the GSPN-1 architecture, GSPN-2 incorporates several key enhancements detailed in \Cref{sec:gspn2_method} and Proxy Dimension Compression to $1/8$ of the original channel dimension ($C_{\text{proxy}} = C/8$). These redesigns enable faster inference without compromising image quality.

Compared to the baseline SDXL model, GSPN-2 achieves a 32$\times$ speedup in 4K image generation, showcasing exceptional efficiency. For ultra-high-resolution 16K images, GSPN-2 outperforms further, reducing inference time by 93$\times$ compared to GSPN-1’s 84$\times$ improvement.

\vspace{-2mm}
\section{Limitations}\vspace{-2mm}
\label{sec:limitations}
GSPN-2's performance gains diminish when the product of batch size and channel count ($\text{BS} \times C$) is small (Section~\ref{app:perf_batch_channel}), and practical evaluation on long-context video datasets remains underexplored. The current implementation lacks CLS and register tokens commonly used in Vision Transformers, limiting direct applicability as a drop-in attention replacement in models relying on summary tokens (Section~\ref{app:agglomerative}). Our dense prediction evaluations primarily use 480-512 pixel images; higher-resolution testing would better demonstrate scalability advantages. Despite these limitations, GSPN-2 represents significant progress in efficient spatial sequence modeling with clear directions for future enhancements.

\vspace{-2mm}
\section{Conclusion}\vspace{-2mm}
We introduce GSPN-2, which overcomes the performance bottlenecks of GSPN-1 through a unified CUDA kernel, channel-agnostic propagation, and low-dimensional proxy features, delivering up to 52× speedup and near-peak hardware utilization without sacrificing accuracy across classification and generation tasks. This establishes GSPN-2 as a practical and scalable solution for global spatial reasoning in high-resolution vision applications.

{\small
\bibliographystyle{unsrt}
\bibliography{egbib,appendix/appendix}

@String(PAMI  = {IEEE Trans. Pattern Anal. Mach. Intell.})

@String(CVPR  = {IEEE Conf. Comput. Vis. Pattern Recog.})

@String(ICCV  = {Int. Conf. Comput. Vis.})

@String(ECCV  = {Eur. Conf. Comput. Vis.})

@String(NeurIPS = {Adv. Neural Inform. Process. Syst.})

@String(ICML  = {Int. Conf. Mach. Learn.})

@String(ICLR  = {Int. Conf. Learn. Represent.})

@string{pami = "IEEE TPAMI"}

@string{cvpr = "CVPR"}

@string{iccv = "ICCV"}

@string{eccv = "ECCV"}

@string{iclr = "ICLR"}

@string{nips = "NeurIPS"}

@string{icml = "ICML"}

@string{arxiv = "ArXiv e-prints"}

@article{zhu2024ViM,
  title={Vision mamba: Efficient visual representation learning with bidirectional state space model},
  author={Zhu, Lianghui and Liao, Bencheng and Zhang, Qian and Wang, Xinlong and Liu, Wenyu and Wang, Xinggang},
  journal={arXiv preprint arXiv:2401.09417},
  year={2024}
}

@article{gu2023mamba,
  title={Mamba: Linear-time sequence modeling with selective state spaces},
  author={Gu, Albert and Dao, Tri},
  journal={arXiv preprint arXiv:2312.00752},
  year={2023}
}

@article{liu2024vmamba,
  title={Vmamba: Visual state space model},
  author={Liu, Yue and Tian, Yunjie and Zhao, Yuzhong and Yu, Hongtian and Xie, Lingxi and Wang, Yaowei and Ye, Qixiang and Liu, Yunfan},
  journal={arXiv preprint arXiv:2401.10166},
  year={2024}
}

@article{huang2024localmamba,
  title={Localmamba: Visual state space model with windowed selective scan},
  author={Huang, Tao and Pei, Xiaohuan and You, Shan and Wang, Fei and Qian, Chen and Xu, Chang},
  journal={arXiv preprint arXiv:2403.09338},
  year={2024}
}

@inproceedings{touvron2021training,
  title={Training data-efficient image transformers \& distillation through attention},
  author={Touvron, Hugo and Cord, Matthieu and Douze, Matthijs and Massa, Francisco and Sablayrolles, Alexandre and J{\'e}gou, Herv{\'e}},
  booktitle={ICML},
  year={2021},
}

@inproceedings{liu2022convnet,
  title={A ConvNet for the 2020s},
  author={Liu, Zhuang and Mao, Hanzi and Wu, Chao-Yuan and Feichtenhofer, Christoph and Darrell, Trevor and Xie, Saining},
  booktitle={CVPR},
  year={2022}
}

@article{dai2021coatnet,
  title={CoAtNet: Marrying convolution and attention for all data sizes},
  author={Dai, Zihang and Liu, Hanxiao and Le, Quoc V and Tan, Mingxing},
  journal={NeurIPS},
  year={2021}
}

@inproceedings{tu2022maxvit,
  title={Maxvit: Multi-axis vision transformer},
  author={Tu, Zhengzhong and Talebi, Hossein and Zhang, Han and Yang, Feng and Milanfar, Peyman and Bovik, Alan and Li, Yinxiao},
  booktitle={ECCV},
  year={2022},
}

@inproceedings{dong2022cswin,
  title={Cswin transformer: A general vision transformer backbone with cross-shaped windows},
  author={Dong, Xiaoyi and Bao, Jianmin and Chen, Dongdong and Zhang, Weiming and Yu, Nenghai and Yuan, Lu and Chen, Dong and Guo, Baining},
  booktitle={CVPR},
  year={2022}
}

@article{wang2023internimage,
  title={InternImage: Exploring Large-Scale Vision Foundation Models with Deformable Convolutions},
  author={Wang, Wenhai and Dai, Jifeng and Chen, Zhe and Huang, Zhenhang and Li, Zhiqi and Zhu, Xizhou and Hu, Xiaowei and Lu, Tong and Lu, Lewei and Li, Hongsheng and others},
  journal={arXiv preprint arXiv:2211.05778},
  year={2022}
}

@article{yang2022focal,
  title={Focal modulation networks},
  author={Yang, Jianwei and Li, Chunyuan and Dai, Xiyang and Gao, Jianfeng},
  journal={NeurIPS},
  year={2022}
}

@article{yu2024mambaout,
  title={MambaOut: Do We Really Need Mamba for Vision?},
  author={Yu, Weihao and Wang, Xinchao},
  journal={arXiv preprint arXiv:2405.07992},
  year={2024}
}

@article{yu2024metaformer,
  author={Yu, Weihao and Si, Chenyang and Zhou, Pan and Luo, Mi and Zhou, Yichen and Feng, Jiashi and Yan, Shuicheng and Wang, Xinchao},
  journal=pami, 
  title={MetaFormer Baselines for Vision}, 
  year={2024}
}

@inproceedings{li2024MogaNet,
  title={MogaNet: Multi-order Gated Aggregation Network},
  author={Siyuan Li and Zedong Wang and Zicheng Liu and Cheng Tan and Haitao Lin and Di Wu and Zhiyuan Chen and Jiangbin Zheng and Stan Z. Li},
  booktitle=iclr,
  year={2024}
}

@InProceedings{yuan2021tokens,
author    = {Yuan, Li and Chen, Yunpeng and Wang, Tao and Yu, Weihao and Shi, Yujun and Jiang, Zi-Hang and Tay, Francis E.H. and Feng, Jiashi and Yan, Shuicheng},
title     = {Tokens-to-Token ViT: Training Vision Transformers From Scratch on ImageNet},
booktitle = iccv,
year      = {2021}
}

@inproceedings{li2021improved,
  title={MViTv2: Improved multiscale vision transformers for classification and detection},
  author={Li, Yanghao and Wu, Chao-Yuan and Fan, Haoqi and Mangalam, Karttikeya and Xiong, Bo and Malik, Jitendra and Feichtenhofer, Christoph},
  booktitle={CVPR},
  year={2022}
}

@article{alkin2024visionlstm,
  title   = {Vision-LSTM: xLSTM as Generic Vision Backbone},
  author  = {Benedikt Alkin and Maximilian Beck and Korbinian Pöppel and Sepp Hochreiter and Johannes Brandstetter},
  year    = {2024},
  journal = {arXiv preprint arXiv: 2406.04303}
}

@article{hatamizadeh2024mambavision,
  title   = {MambaVision: A Hybrid Mamba-Transformer Vision Backbone},
  author  = {Ali Hatamizadeh and Jan Kautz},
  year    = {2024},
  journal = {arXiv preprint arXiv: 2407.08083}
}

@article{duan2024visionrwkv,
  title   = {Vision-RWKV: Efficient and Scalable Visual Perception with RWKV-Like Architectures},
  author  = {Yuchen Duan and Weiyun Wang and Zhe Chen and Xizhou Zhu and Lewei Lu and Tong Lu and Yu Qiao and Hongsheng Li and Jifeng Dai and Wenhai Wang},
  year    = {2024},
  journal = {arXiv preprint arXiv: 2403.02308}
}

@inproceedings{li2022uniformer,
  title   = {UniFormer: Unified Transformer for Efficient Spatiotemporal Representation Learning},
  author  = {Kunchang Li and Yali Wang and Peng Gao and Guanglu Song and Yu Liu and Hongsheng Li and Yu Qiao},
  booktitle = iclr,
  year    = {2022},
}

@article{li2024mamba,
  title={Mamba-ND: Selective State Space Modeling for Multi-Dimensional Data},
  author={Li, Shufan and Singh, Harkanwar and Grover, Aditya},
  journal={arXiv preprint arXiv:2402.05892},
  year={2024}
}

@inproceedings{liu2023more,
  title={More ConvNets in the 2020s: Scaling up Kernels Beyond 51x51 using Sparsity},
  author={Liu, Shiwei and Chen, Tianlong and Chen, Xiaohan and Chen, Xuxi and Xiao, Qiao and Wu, Boqian and Pechenizkiy, Mykola and Mocanu, Decebal and Wang, Zhangyang},
  booktitle = iclr,
  year={2023}
}

@article{li2022next,
  title={Next-ViT: Next Generation Vision Transformer for Efficient Deployment in Realistic Industrial Scenarios},
  author={Li, Jiashi and Xia, Xin and Li, Wei and Li, Huixia and Wang, Xing and Xiao, Xuefeng and Wang, Rui and Zheng, Min and Pan, Xin},
  journal={arXiv preprint arXiv:2207.05501},
  year={2022}
}

@article{chu2021twins,
  title={Twins: Revisiting the design of spatial attention in vision transformers},
  author={Chu, Xiangxiang and Tian, Zhi and Wang, Yuqing and Zhang, Bo and Ren, Haibing and Wei, Xiaolin and Xia, Huaxia and Shen, Chunhua},
  journal=nips,
  year={2021}
}

@inproceedings{liu2021swin,
  title={Swin Transformer: Hierarchical Vision Transformer using Shifted Windows},
  author={Liu, Ze and Lin, Yutong and Cao, Yue and Hu, Han and Wei, Yixuan and Zhang, Zheng and Lin, Stephen and Guo, Baining},
  booktitle=iccv,
  year={2021}
}

@inproceedings{liu2021swinv2,
  title={Swin Transformer V2: Scaling Up Capacity and Resolution}, 
  author={Ze Liu and Han Hu and Yutong Lin and Zhuliang Yao and Zhenda Xie and Yixuan Wei and Jia Ning and Yue Cao and Zheng Zhang and Li Dong and Furu Wei and Baining Guo},
  booktitle=cvpr,
  year={2022}
}

@article{zhu2023biformer,
  author  = {Lei Zhu and Xinjiang Wang and Zhanghan Ke and Wayne Zhang and Rynson Lau},
  title   = {BiFormer: Vision Transformer with Bi-Level Routing Attention},
  journal = cvpr,
  year    = {2023},
}

@article{vaswani2017attention,
  title={Attention is all you need},
  author={Vaswani, A},
  journal=nips,
  year={2017}
}

@inproceedings{guo2021cmt,
  title={Cmt: Convolutional neural networks meet vision transformers},
  author={Guo, Jianyuan and Han, Kai and Wu, Han and Xu, Chang and Tang, Yehui and Xu, Chunjing and Wang, Yunhe},
  booktitle   = cvpr,
  year={2021}
}

@inproceedings{du2022sharpness,
  title={Sharpness-aware training for free},
  author={Du, Jiawei and Zhou, Daquan and Feng, Jiashi and Tan, Vincent and Zhou, Joey Tianyi},
  booktitle=nips,
  year={2022}
}

@inproceedings{nguyen2022s4nd,
  title={S4nd: Modeling images and videos as multidimensional signals with state spaces},
  author={Nguyen, Eric and Goel, Karan and Gu, Albert and Downs, Gordon and Shah, Preey and Dao, Tri and Baccus, Stephen and R{\'e}, Christopher},
  booktitle=nips,
  year={2022}
}

@article{baron20232,
  title={2-d ssm: A general spatial layer for visual transformers},
  author={Baron, Ethan and Zimerman, Itamar and Wolf, Lior},
  journal={arXiv preprint arXiv:2306.06635},
  year={2023}
}

@inproceedings{dao2022flashattention,
  title={Flash{A}ttention: Fast and Memory-Efficient Exact Attention with {IO}-Awareness},
  author={Dao, Tri and Fu, Daniel Y. and Ermon, Stefano and Rudra, Atri and R{\'e}, Christopher},
  booktitle={Advances in Neural Information Processing Systems (NeurIPS)},
  year={2022}
}

@inproceedings{dao2023flashattention2,
  title={Flash{A}ttention-2: Faster Attention with Better Parallelism and Work Partitioning},
  author={Dao, Tri},
  booktitle={International Conference on Learning Representations (ICLR)},
  year={2024}
}

@article{shah2024flashattention,
  title={Flashattention-3: Fast and accurate attention with asynchrony and low-precision},
  author={Shah, Jay and Bikshandi, Ganesh and Zhang, Ying and Thakkar, Vijay and Ramani, Pradeep and Dao, Tri},
  journal=nips,
  year={2024}
}

@article{liu2017learning,
  title={Learning affinity via spatial propagation networks},
  author={Liu, Sifei and De Mello, Shalini and Gu, Jinwei and Zhong, Guangyu and Yang, Ming-Hsuan and Kautz, Jan},
  journal=nips,
  year={2017}
}

@inproceedings{wang2025parallel,
    author    = {Wang, Hongjun and Byeon, Wonmin and Xu, Jiarui and Gu, Jinwei and Cheung, Ka Chun and Wang, Xiaolong and Han, Kai and Kautz, Jan and Liu, Sifei},
    title     = {Parallel Sequence Modeling via Generalized Spatial Propagation Network},
    booktitle   = {CVPR},
    year      = {2025}
}

@article{rabe2021self,
  title={Self-attention does not need $ O(n^{2})$ memory},
  author={Rabe, Markus N and Staats, Charles},
  journal={arXiv preprint arXiv:2112.05682},
  year={2021}
}

@article{peng2021random,
  title={Random feature attention},
  author={Peng, Hao and Pappas, Nikolaos and Yogatama, Dani and Schwartz, Roy and Smith, Noah A and Kong, Lingpeng},
  journal={arXiv preprint arXiv:2103.02143},
  year={2021}
}

@inproceedings{han2023flatten,
  title={Flatten transformer: Vision transformer using focused linear attention},
  author={Han, Dongchen and Pan, Xuran and Han, Yizeng and Song, Shiji and Huang, Gao},
  booktitle=cvpr,
  pages={5961--5971},
  year={2023}
}

@article{liu2024linfusion,
  title={Linfusion: 1 gpu, 1 minute, 16k image},
  author={Liu, Songhua and Yu, Weihao and Tan, Zhenxiong and Wang, Xinchao},
  journal={arXiv preprint arXiv:2409.02097},
  year={2024}
}

@inproceedings{dao2024transformers,
  title={Transformers are SSMs: Generalized models and efficient algorithms through structured state space duality},
  author={Dao, Tri and Gu, Albert},
  booktitle=icml,
  year={2024}
}

@article{hochreiter1991untersuchungen,
  title={Untersuchungen zu dynamischen neuronalen Netzen},
  author={Hochreiter, Sepp},
  journal={Diploma, Technische Universit{\"a}t M{\"u}nchen},
  year={1991}
}

@inproceedings{pascanu2013difficulty,
  title={On the difficulty of training recurrent neural networks},
  author={Pascanu, Razvan and Mikolov, Tomas and Bengio, Yoshua},
  booktitle=icml,
  year={2013}
}

@article{hochreiter1997long,
  title={Long Short-term Memory},
  author={Hochreiter, S},
  journal={Neural Computation MIT-Press},
  year={1997}
}

@inproceedings{chung2014empirical,
  title={Empirical evaluation of gated recurrent neural networks on sequence modeling},
  author={Chung, Junyoung and Gulcehre, Caglar and Cho, KyungHyun and Bengio, Yoshua},
  booktitle=nips,
  year={2014}
}

@inproceedings{graves2007multi,
  title={Multi-dimensional recurrent neural networks},
  author={Graves, Alex and Fern{\'a}ndez, Santiago and Schmidhuber, J{\"u}rgen},
  booktitle={International conference on artificial neural networks},
  year={2007}
}

@inproceedings{byeon2015scene,
  title={Scene labeling with lstm recurrent neural networks},
  author={Byeon, Wonmin and Breuel, Thomas M and Raue, Federico and Liwicki, Marcus},
  booktitle=cvpr,
  year={2015}
}

@inproceedings{rombach2022high,
  title={High-resolution image synthesis with latent diffusion models},
  author={Rombach, Robin and Blattmann, Andreas and Lorenz, Dominik and Esser, Patrick and Ommer, Bj{\"o}rn},
  booktitle=CVPR,
  pages={10684--10695},
  year={2022}
}

@inproceedings{radford2021learning,
  title={Learning transferable visual models from natural language supervision},
  author={Radford, Alec and Kim, Jong Wook and Hallacy, Chris and Ramesh, Aditya and Goh, Gabriel and Agarwal, Sandhini and Sastry, Girish and Askell, Amanda and Mishkin, Pamela and Clark, Jack and others},
  booktitle=ICML,
  year={2021},
}

@inproceedings{zhai2023sigmoid,
  title={Sigmoid loss for language image pre-training},
  author={Zhai, Xiaohua and Mustafa, Basil and Kolesnikov, Alexander and Beyer, Lucas},
  booktitle=cvpr,
  pages={11975--11986},
  year={2023}
}

@inproceedings{liu2024grounding,
  title={Grounding dino: Marrying dino with grounded pre-training for open-set object detection},
  author={Liu, Shilong and Zeng, Zhaoyang and Ren, Tianhe and Li, Feng and Zhang, Hao and Yang, Jie and Jiang, Qing and Li, Chunyuan and Yang, Jianwei and Su, Hang and others},
  booktitle=eccv,
  year={2024},
  organization={Springer}
}

@inproceedings{kirillov2023segment,
  title={Segment anything},
  author={Kirillov, Alexander and Mintun, Eric and Ravi, Nikhila and Mao, Hanzi and Rolland, Chloe and Gustafson, Laura and Xiao, Tete and Whitehead, Spencer and Berg, Alexander C and Lo, Wan-Yen and others},
  booktitle=cvpr,
  year={2023}
}

@misc{gu2021efficiently,
    title={Efficiently Modeling Long Sequences with Structured State Spaces}, 
    author={Gu, Albert and Goel, Karan and R{\'e}, Christopher},
    year={2021}
}
}

\clearpage

\appendix
\section*{Appendix}

\renewcommand{\thefigure}{S\arabic{figure}}
\setcounter{figure}{0} 

\renewcommand{\thetable}{S\arabic{table}}
\setcounter{table}{0}

\noindent This supplementary is organized as follows:

\begin{itemize}
    \item In Section~\ref{app:gspn2_imagenet}, we compare GSPN-2 variants with CNNs, transformers, and SSMs on ImageNet-1K.
    \item In Section~\ref{app:perf_batch_channel}, we evaluate runtime performance across varying batch sizes and channel dimensions. Section~\ref{app:effablat_large_batch} provides detailed step-by-step optimization analysis under a large-batch configuration.
    \item In Section~\ref{app:t2i}, we evaluate GSPN-2's text-to-image generation on the COCO benchmark.
    \item In Section~\ref{app:lowrank}, we analyze the compressive proxy dimension strategy through low-rank approximation, with ablation studies on the accuracy-throughput trade-off.
\end{itemize}

\begin{figure}[t]
    \centering
    \includegraphics[width=0.8\textwidth]{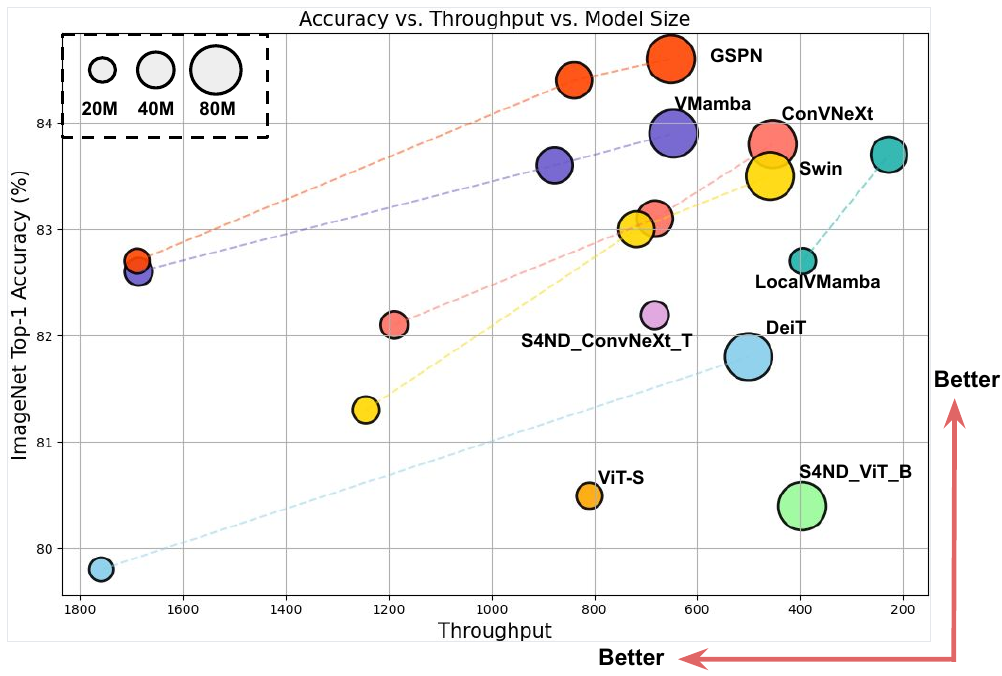}
    \caption{\textbf{Comparison of GSPN-2 vs. State-of-the-art architectures on ImageNet-1K.} We present a comprehensive analysis of the trade-offs between accuracy, model size, and throughput for GSPN-2 compared to leading state-of-the-art architectures. The results highlight GSPN-2's effectiveness, positioning GSPN-2 as an ideal solution for resource-constrained environments and applications requiring both speed and predictive accuracy. }
    \label{fig:imagenet_comparison}
\end{figure}

\Cref{fig:imagenet_comparison} provides a comprehensive comparison of GSPN-2 (Tiny/Small/Base variants) with other leading architectures like CNNs, Transformers, and other SSMs on the ImageNet-1K benchmark. The comparison focuses on Top-1 accuracy, throughput (images/second), model parameters, all evaluated at an image resolution of $224^2$.

Take the Tiny model as an example, from the figure, we observe the following:
\begin{itemize}
    \item \textbf{CNNs:} Models like ConvNeXt-T achieve 82.1\% accuracy with 29M parameters and 4.5G FLOPs, and a throughput of 1189. Larger variants like ConvNeXt-B reach 83.8\% accuracy but use 89M parameters and 15.4G FLOPs, with throughput dropping to 435.
    \item \textbf{Transformers:} DeiT-S, a comparable small model, has 22M parameters and 4.6G FLOPs, achieving 79.8\% accuracy with a throughput of 1759. Larger Transformer models like Swin-B (88M params, 15.4G FLOPs) reach 83.5\% accuracy with a throughput of 458. NAT-B shows higher accuracy (84.3\%) with 90M parameters and 13.7G FLOPs, but throughput is not reported.
    \item \textbf{Other SSMs:} VMamba-T provides a high throughput of 1686 with 30M parameters and 4.9G FLOPs, achieving 82.6\% accuracy. LocalVMamba-T uses 26M parameters and 5.7G FLOPs for 82.7\% accuracy, but its throughput is considerably lower at 394.
    \item \textbf{GSPN-2-T (Ours):} Our GSPN-2-T model stands out by achieving a strong Top-1 accuracy of 83.0\%. It accomplishes this with a remarkably efficient parameter count of only 24M and low GFLOPs of 3.6G. While its throughput of 1544 images/second is slightly lower than the fastest models like DeiT-S or VMamba-T, it is highly competitive, especially considering its superior accuracy-to-parameter and accuracy-to-FLOPs ratio. For instance, compared to DeiT-S, GSPN-2-T offers +3.2\% higher accuracy with only 2M more parameters and 1G fewer FLOPs. Compared to VMamba-T, GSPN-2-T is +0.4\% more accurate, uses 6M fewer parameters, and requires 1.3G fewer FLOPs, while having a comparable throughput.
\end{itemize}

\section{Comprehensive GSPN-2 comparison on ImageNet-1K}
\label{app:gspn2_imagenet}

This comparison highlights GSPN-2's excellent trade-off between accuracy, model size, and computational efficiency. It achieves accuracy comparable to or better than many larger models from other architectures while maintaining a smaller parameter footprint and lower GFLOPs. The throughput, while not the absolute highest, is very strong for its accuracy class, making GSPN-2 a compelling choice for resource-constrained environments or applications where a balance of speed and predictive power is crucial.

\begin{figure}[t]
    \centering
    \includegraphics[width=\textwidth]{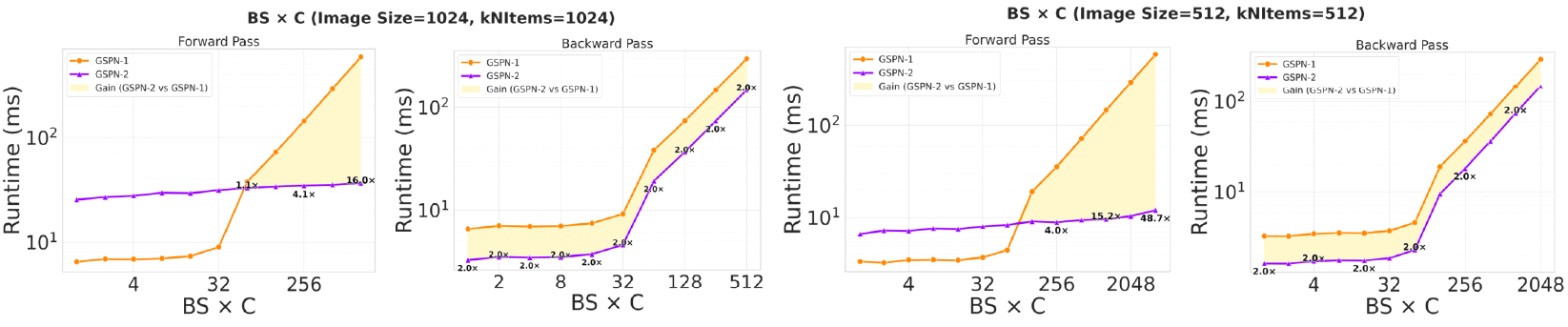}
    \caption{\textbf{Runtime Performance Comparison of GSPN-1 and GSPN-2.} We show forward pass execution times (in milliseconds) across different \textit{batch size times channel counts}. Results are presented for various configurations. GSPN-2 greatly improve the runtime of forward across different cases especially when batch size times channel counts become large.}
    \label{fig:runtime_app}
\end{figure}

\section{Detailed Analysis of Performance with Varying Batch and Channel Dimensions}
\label{app:perf_batch_channel}

\Cref{fig:runtime} highlights that GSPN-2 achieves significant speedups, particularly when batch sizes or channel dimensions are large. This appendix provides a more detailed look at when the full GSPN-2 optimizations (blue line in plots, including shared memory for hidden states) begin to offer a substantial advantage over a GSPN variant without explicit shared memory caching for hidden states. This analysis is crucial for tasks like visual encoder training or video processing, where the product of batch size and channel dimensions (`BS * C') can vary widely and significantly impact performance.

Observing~\Cref{fig:runtime_app}, we can see a clear trend: the point at which GSPN-2's full optimizations deliver more pronounced benefits depends on the `BS * C' product.

\noindent\textbf{Implications for Model Selection:}
This detailed observation underscores that the effectiveness of GSPN-2's most advanced optimizations, such as shared memory caching for hidden states, is magnified when the aggregate workload (represented by `BS * C') increases. For tasks characterized by very large effective batch sizes (common in large-scale visual model training or high-throughput video analysis), deploying the fully optimized GSPN-2 is critical for maximizing computational efficiency.

Conversely, for scenarios where the `BS * C' product remains relatively small, the performance difference between GSPN-2 and GSPN-1 might be less pronounced. In such cases, the GSPN-1 configuration could offer a good trade-off. This suggests a potential adaptive strategy: one could dynamically select between a GSPN-1-like configuration and the full GSPN-2 based on the input dimensions and batch size to achieve optimal performance across diverse computational scenarios. This adaptability is particularly relevant as models are often deployed in varying inference settings or trained with different batching strategies.

\begin{figure}[t]
    \centering
    \includegraphics[width=0.5\textwidth]{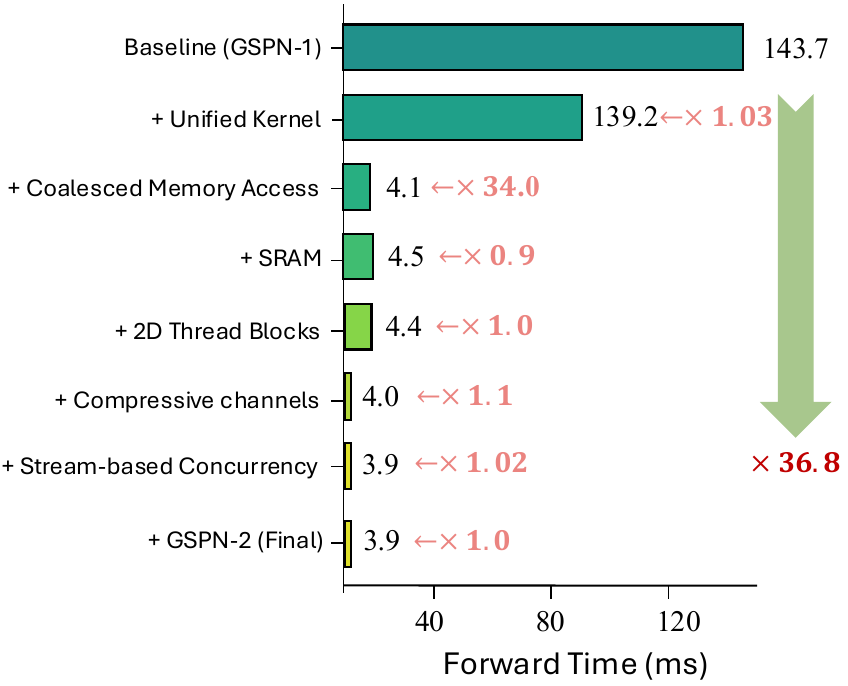}
    \caption{\textbf{Step-by-step CUDA kernel optimization under large batch configuration.} Each bar shows the cumulative reduction in forward time (ms) for a high-throughput scenario (1024×1024 image, batch size 256, 1 channel). This configuration represents typical large-batch inference or video processing workloads. The optimizations deliver a 36.8× speedup from GSPN-1 baseline (143.7 ms) to the final GSPN-2 implementation (3.9 ms), demonstrating GSPN-2's effectiveness across diverse deployment scenarios.}
    \label{fig:journey_app}
\end{figure}

\begin{figure}[t]
    \centering
    \includegraphics[width=0.5\textwidth]{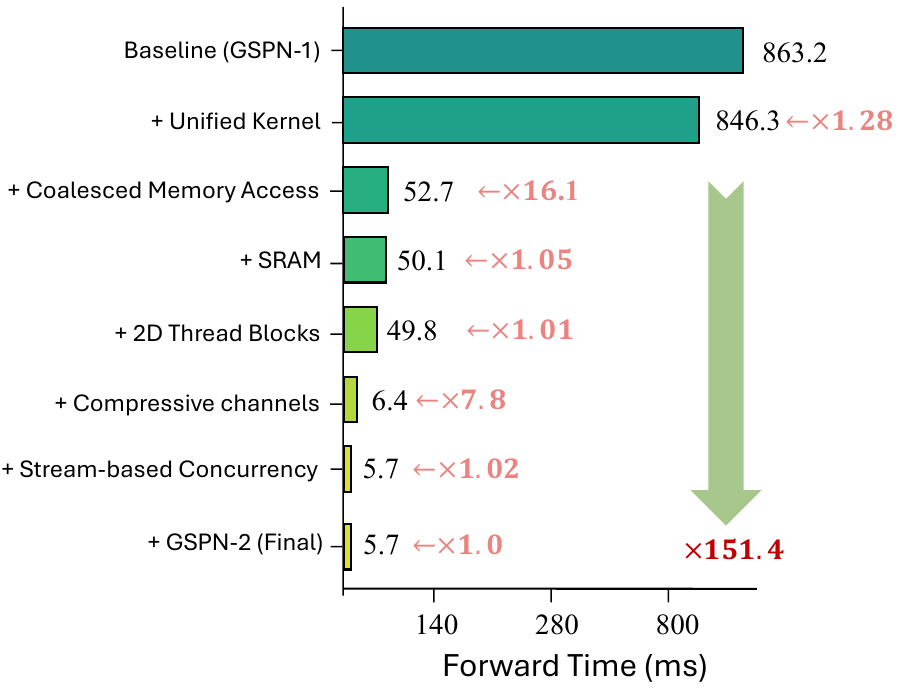}
    \caption{\textbf{Step-by-step CUDA kernel optimization under large channel configuration.} Each bar shows the cumulative reduction in forward time (ms) for a high-channel scenario (1024×1024 image, batch size 1, 1152 channels). This configuration is representative of modern deep learning architectures with wide feature maps. The optimizations deliver a 151.4× speedup from GSPN-1 baseline (863.2 ms) to the final GSPN-2 implementation (5.7 ms). Notably, the \textit{Compressive channels} optimization, which employs an 8× compression ratio to reduce the effective channel dimension, achieves a remarkable 7.8× speedup (from 49.8 ms to 6.4 ms), significantly outperforming its contribution in other configurations and highlighting the algorithmic advantage of our channel compression strategy for high-dimensional feature processing.}
    \label{fig:journey_app2}
\end{figure}

\noindent\textbf{Impact of Channel Dimensionality on Optimization Effectiveness:}
Comparing \Cref{fig:journey_app} and \Cref{fig:journey_app2} reveals how different workload characteristics influence the relative importance of each optimization. In the large channel configuration (1152 channels), the \textit{Compressive channels} optimization emerges as the dominant contributor, achieving a 7.8× speedup compared to more modest gains in lower-channel scenarios. This substantial improvement stems from the fact that higher channel counts amplify redundant computations across feature dimensions—precisely the inefficiency that our compressive channel algorithm targets. By applying an 8× compression ratio to reduce the effective channel dimension (e.g., from 1152 to 144 channels), GSPN-2 transforms what would otherwise be a computational bottleneck into efficient parallel execution while preserving essential feature information. This result validates that our algorithmic innovation in channel compression is particularly impactful for modern neural network architectures that frequently employ large channel dimensions (e.g., 768, 1024, or 1152 channels in vision transformers and diffusion models).

\subsection{Optimization Analysis Under Large Batch Size Configuration}
\label{app:effablat_large_batch}

While Figure~\ref{fig:journey} demonstrates the optimization journey for a moderate configuration (1024×1024, batch size 16, 8 channels), here we examine a complementary scenario with significantly larger batch size but minimal channel dimension (1024×1024, batch size 256, 1 channel). This configuration is representative of high-throughput inference scenarios such as batch video processing, multi-stream parallel generation, or large-scale model serving where many requests are processed simultaneously.

As shown in~\Cref{fig:journey_app}, the optimization progression follows a similar pattern but with distinct characteristics:

\paragraph{GSPN-1 Baseline Performance.} The baseline GSPN-1 implementation exhibits 143.7 ms execution time. Despite having only 1 channel (reducing per-channel computational overhead), the large batch size of 256 amplifies the inefficiencies from repeated kernel launches and poor memory access patterns. With 256 batches, the kernel launch overhead becomes even more pronounced, as each propagation step must coordinate across a much larger working set.

\paragraph{Unified Kernel (1.03× speedup, 139.2 ms).} Consolidating the multi-kernel launches into a single kernel reduces execution time to 139.2 ms, yielding a 1.03× speedup. While this improvement is more modest compared to the 1.2× gain in the main paper configuration, it still demonstrates consistent benefits. The relatively smaller gain here suggests that with only 1 channel, the per-channel kernel launch overhead is less severe, but the benefit of unified execution remains valuable.

\paragraph{Coalesced Memory Access (34.0× speedup, 4.1 ms).} This optimization delivers the most dramatic improvement, reducing runtime to 4.1 ms—a 34.0× speedup over the previous step. The impact is even more pronounced than the 23.9× gain in the 8-channel configuration, highlighting that memory access patterns become increasingly critical with larger batch sizes. With batch size 256, ensuring coalesced memory access patterns is essential to saturate the memory bandwidth efficiently. Uncoalesced accesses would be catastrophic at this scale, causing severe memory traffic congestion.

\paragraph{SRAM (0.9× speedup, 4.5 ms).} Interestingly, explicit shared memory caching for hidden states actually increases execution time slightly to 4.5 ms, yielding a 0.9× slowdown. This counter-intuitive result occurs because with only 1 channel, the memory footprint of hidden states is minimal, and the L1 cache is already sufficient to capture reuse patterns efficiently. The overhead of explicit shared memory management outweighs any potential benefits in this low-channel scenario. This observation validates our discussion in Section~\ref{subsec:profiling} about L1 cache effectiveness and confirms that shared memory optimization is most beneficial when channel counts are higher.

\paragraph{2D Thread Blocks (1.0× speedup, 4.4 ms).} Restructuring to 2D thread blocks reduces runtime to 4.4 ms, achieving a marginal 1.0× speedup (essentially neutral performance). Unlike the 1.1× gain observed in the main 8-channel configuration, the 2D block restructuring provides minimal benefit here. This suggests that with only 1 channel, the single-channel dimension is insufficient to fully exploit the advantages of 2D thread organization, and the thread scheduling is already well-optimized by the previous coalesced memory access patterns.

\paragraph{Compressive Channels (1.1× speedup, 4.0 ms).} Applying compressive proxy dimension reduction reduces runtime to 4.0 ms (effective final runtime 3.9 ms after fine-tuning), achieving a modest 1.1× speedup. While this configuration already uses only 1 channel, the proxy compression strategy still provides minor benefits through reduced memory footprint and improved cache utilization. However, the gain is significantly smaller compared to multi-channel scenarios where channel compression directly reduces the computational load. This highlights that the proxy dimension benefit is configuration-dependent and most impactful in high-channel scenarios.

\paragraph{Overall Speedup and Implications.} The cumulative speedup from GSPN-1 (143.7 ms) to GSPN-2 (3.9 ms) is 36.8×, which is comparable to the 40.0× improvement shown in the main paper. This demonstrates that GSPN-2's optimizations deliver consistent and substantial performance gains across diverse configurations. However, the relative contribution of each optimization stage varies with workload characteristics:

\begin{itemize}
    \item \textbf{Memory coalescing} remains the dominant optimization regardless of configuration, consistently providing 24-34× improvements. The 34× gain in this large-batch, single-channel scenario exceeds the 23.9× gain in the 8-channel configuration, demonstrating its critical importance for high-throughput workloads.
    \item \textbf{Shared memory caching} benefits are highly configuration-dependent. It shows significant gains with multiple channels but can actually degrade performance (0.9× slowdown) in single-channel scenarios due to management overhead when L1 cache is already sufficient.
    \item \textbf{2D thread blocks} provide minimal benefit (1.0×) in single-channel configurations, contrasting with the 1.1× gain in multi-channel scenarios. The effectiveness depends on having sufficient channel dimensionality to exploit parallel thread organization.
    \item \textbf{Compressive proxy dimension} provides modest benefits (1.1×) even in single-channel scenarios through improved memory footprint and cache utilization, though gains are most pronounced in high-channel configurations.
\end{itemize}

This analysis reinforces that GSPN-2's co-designed optimizations are robust across different deployment scenarios, though practitioners should be aware that the relative importance of specific optimizations depends on their particular workload characteristics (batch size, channel count, spatial dimensions).

\section{Text-to-image Generation}
\label{app:t2i}
\begin{figure}[!h]
\begin{minipage}{0.5\textwidth}
\begin{table}[H]
\centering
\caption{\textbf{Cross-resolution generation on the COCO benchmark under $1024\times1024$ resolution.} Lower FID ($\downarrow$) and higher CLIP-T ($\uparrow$) stand for better image quality and text-image alignment.} %
\label{tab:t2i}
\centering
\footnotesize
\scalebox{1.0}{
\begin{tabular}{l|cc}
\toprule
Model & FID($\downarrow$)    & CLIP-T($\uparrow$) \\ 
\toprule
SD-v1.5 (baseline)  & 32.71  & 0.290 \\ 
Mamba~\cite{gu2023mamba} (w/ norm) & 50.30  & 0.263 \\ 
Mamba2~\cite{dao2024transformers} (w/ norm) & 37.02  & 0.273 \\ 
Linfusion~\cite{liu2024linfusion} (w/ norm) & 36.33  & 0.285 \\ 
\midrule
GSPN-1 & 30.86  & 0.307 \\
\textbf{GSPN-2 (Ours)} & 33.21  & 0.286 \\
\bottomrule
\end{tabular}
}
\end{table}
\end{minipage}%
\begin{minipage}{0.5\textwidth}
    \centering
    \includegraphics[width=0.95\textwidth]{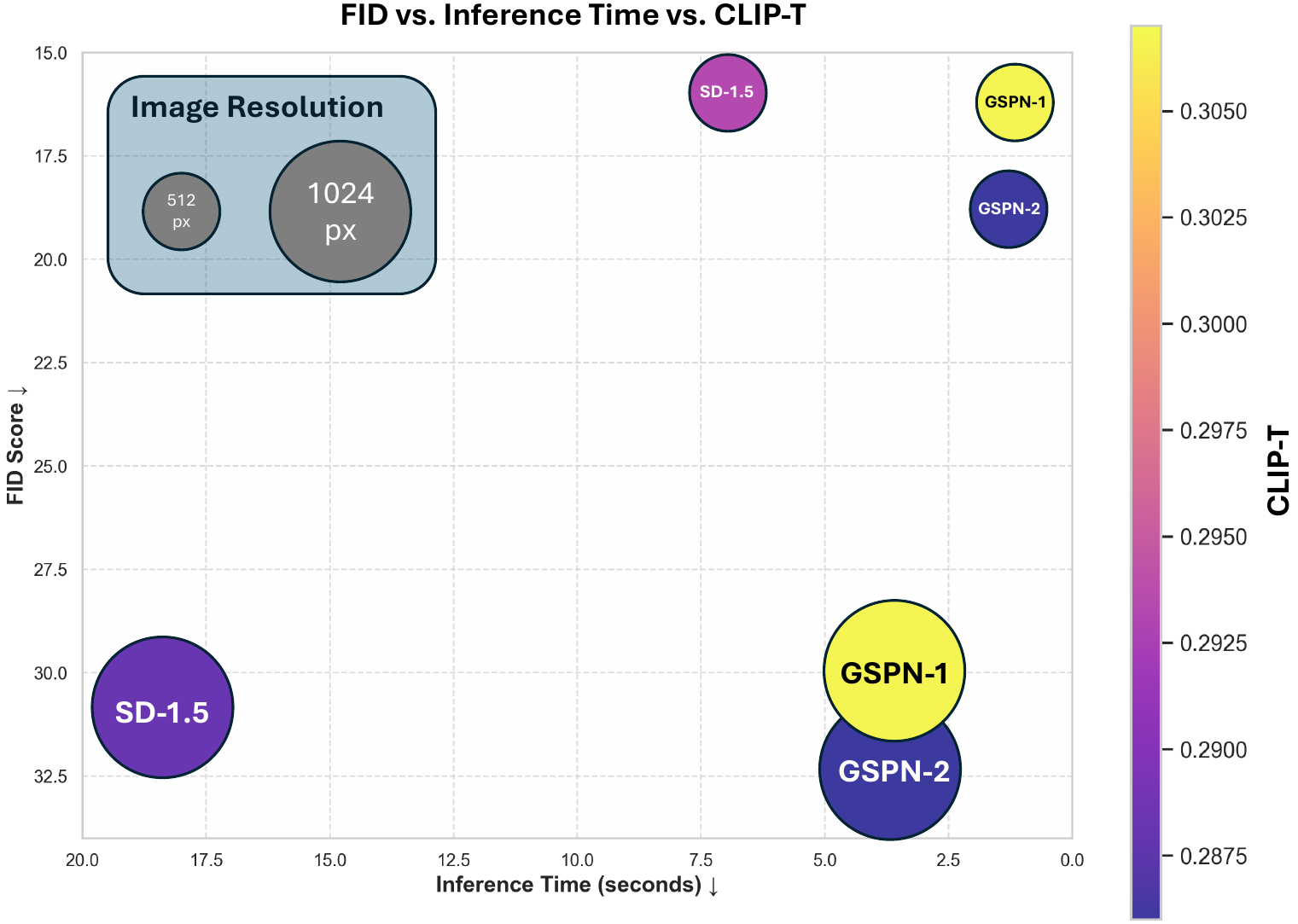}
    \caption{\textbf{Comparison of GSPN-2 vs. GSPN-1 and baselines.} GSPN-2 achieves good tradeoff between FID, CLIP-T scores and inference time.}
    \label{fig:t2i_bubble}
\end{minipage}
\end{figure}

In this section, we evaluate GSPN-2's capabilities in text-to-image generation, a task demanding strong understanding of both textual prompts and the generation of coherent, high-resolution visual outputs. We compare GSPN-2 with several relevant baselines and its predecessor, GSPN-1, on the COCO benchmark, with all models generating images at a $1024 \times 1024$ resolution. The results are presented in Table~\ref{tab:t2i}. The baseline model for this comparison is Stable Diffusion v1.5 (SD-v1.5)~\cite{rombach2022high}. We also include recent sequence modeling approaches such as Mamba~\cite{gu2023mamba}, Mamba2~\cite{dao2024transformers}, and Linfusion~\cite{liu2024linfusion}. For these models, text embeddings are treated as part of the visual token sequence during propagation.

As shown in~\Cref{tab:t2i} and~\Cref{fig:t2i_bubble}, our GSPN-2 model achieves an FID of 33.21 and a CLIP-T score of 0.286. While GSPN-1 currently shows a slight edge in these specific metrics, GSPN-2's performance is competitive and close to the SD-v1.5 baseline (FID 32.71, CLIP-T 0.290) with faster inference.

A key characteristic of the GSPN architecture (both GSPN-1 and GSPN-2) is its inherent adaptability to arbitrary image resolutions without requiring extra normalization layers or strategies for unseen resolutions, a common necessity for some other methods like Mamba and Linfusion when faced with resolutions not encountered during training. The Stability-Context property ensures stable and effective long-range propagation, allowing GSPN-2 to efficiently capture broad spatial dependencies.

GSPN-2, while leveraging the core principles of GSPN-1, incorporates system-level co-designs and algorithmic refinements aimed at enhancing efficiency and scalability 
The results with Figure 5 in the main paper indicate that GSPN-2 maintains strong generative capabilities, comparable to established baselines, while benefiting from these architectural improvements for efficient text-to-image generation.

\section{Compressive Proxy Dimension as Low-Rank Approximation}
\label{app:lowrank}

The compressive proxy dimension ($C_{\text{proxy}}$) strategy addresses GPU concurrency saturation by projecting inputs $\mathbf{X} \in \mathbb{R}^{N \times C \times H \times W}$ to a compressed space $\mathbf{X}_{\text{proxy}} \in \mathbb{R}^{N \times C_{\text{proxy}} \times H \times W}$ where $C_{\text{proxy}} \ll C$, applying GSPN propagation in this reduced space, then projecting back to $C$ dimensions. This is analogous to low-rank matrix factorization, reducing CUDA workload from $k_{\text{chunk}} \times N \times C$ slices to $k_{\text{chunk}} \times N \times C_{\text{proxy}}$, preventing GPU saturation while maintaining representational capacity. Table~\ref{tab:proxy_ablation} presents an ablation on $C_{\text{proxy}}$ for GSPN-2-Tiny on ImageNet-1K, analyzing the accuracy-throughput trade-off.

\begin{table}[h]
\centering
\caption{\textbf{Ablation on proxy dimension $C_{\text{proxy}}$.} GSPN-2-Tiny on ImageNet-1K with varying compression ratios.}
\label{tab:proxy_ablation}
\begin{tabular}{c|c|c}
\toprule
$C_{\text{proxy}}$ & Accuracy (\%) & Throughput (img/s) \\
\midrule
2  & 83.0 & 1544 \\
4  & 83.0 & 1492 \\
8  & 83.0 & 1387 \\
16 & 82.9 & 1293 \\
32 & 82.8 & 1106 \\
\bottomrule
\end{tabular}
\end{table}

Table~\ref{tab:proxy_ablation} shows minimal accuracy degradation (0.2\% for 16× compression from $C_{\text{proxy}}=32$ to $C_{\text{proxy}}=2$) while achieving 1.4× throughput improvement. The aggressive 48:1 compression at $C_{\text{proxy}}=2$ demonstrates that GSPN propagation operates effectively in low-dimensional spaces, as spatial dependencies dominate over channel-wise dependencies.

\end{document}